\documentclass[12pt,a4paper]{article}

\usepackage[british]{babel}

\usepackage[a4paper,top=2cm,bottom=2cm,left=2.5cm,right=2.5cm,marginparwidth=1.75cm]{geometry}

\usepackage[numbers]{natbib}
\usepackage{amsmath}
\usepackage{graphicx}
\usepackage[colorlinks=true, allcolors=blue]{hyperref}
\usepackage{hyperref}
\usepackage[title]{appendix}
\usepackage{mathrsfs}
\usepackage{amsfonts}
\usepackage{booktabs} % For \toprule, \midrule, \botrule
\usepackage{caption}  % For \caption
\usepackage{threeparttable} % For table footnotes
\usepackage{algorithm}
\usepackage{algorithmicx}
\usepackage{algpseudocode}
\usepackage{listings}
\usepackage{enumitem}
\usepackage{chngcntr}
\usepackage{booktabs}
\usepackage{lipsum}
\usepackage{subcaption}
\usepackage{authblk}
\usepackage[T1]{fontenc}    % Font encoding
\usepackage{csquotes}       % Include csquotes
\usepackage{diagbox}
\usepackage{subcaption}
\usepackage{graphicx}
\usepackage{float}
\usepackage{caption}
\usepackage{subcaption}
\usepackage{tikz}
\usepackage{amsmath}
\usepackage{color}
\usepackage{gensymb}

\usepackage[symbol]{footmisc}

\usepackage{setspace}
\usepackage{titlesec}
\titleformat{\section} % Redefine section numbering format
  {\normalfont\Large\bfseries}{\thesection.}{1em}{}
  
\usepackage{lineno} % Add the lineno package

\rightlinenumbers % Right-align line numbers

\usepackage{float}   % for customizing caption position
\usepackage{caption} % for customizing caption format


\title{A temporal deep learning framework for calibration of low-cost air quality sensors}

\author[1,*]{Arindam Sengupta}
\author[2]{Tony Bush}
\author[2]{Ben Marner}
\author[1]{Jose Miguel Pérez}
\author[1]{Soledad Le Clainche}

\affil[1]{\small ETSI Aeronáutica y del Espacio, Universidad Politécnica de Madrid, Plaza Cardenal Cisneros, 3, Madrid, 28040, Spain}

\affil[2]{\small Air Quality Consultants Ltd., 3rd Floor, St. Augustine's Court, St. Augustine's Place, Bristol, BS1 4UD, United Kingdom}

\affil[*]{Corresponding author: \texttt{a.sengupta@upm.es} (Arindam Sengupta) \\
Co-authors: \texttt{tonybush@noiseconsultants.co.uk}\\
\texttt{benmarner@aqconsultants.co.uk}\\
\texttt{josemiguel.perez@upm.es}\\
\texttt{soledad.leclainche@upm.es}}

\date{}

\begin{document}
\maketitle

\begin{abstract}

Low-cost air quality sensors (LCS) provide a practical alternative to expensive regulatory-grade instruments, making dense urban monitoring networks possible. Yet their adoption is limited by calibration challenges, including sensor drift, environmental cross-sensitivity, and variability in performance from device to device. This work presents a deep learning framework for calibrating LCS measurements of PM$_{2.5}$, PM$_{10}$, and NO$_2$ using a Long Short-Term Memory (LSTM) network, trained on co-located reference data from the OxAria network in Oxford, UK. Unlike the Random Forest (RF) baseline, which treats each observation independently, the proposed approach captures temporal dependencies and delayed environmental effects through sequence-based learning, achieving higher $R^2$ values across training, validation, and test sets for all three pollutants. A feature set is constructed combining time-lagged parameters, harmonic encodings, and interaction terms to improve generalization on unseen temporal windows. Validation of unseen calibrated values against the Equivalence Spreadsheet Tool 3.1 demonstrates regulatory compliance with expanded uncertainties of 22.11\% for NO$_2$, 12.42\% for PM$_{10}$, and 9.1\% for PM$_{2.5}$.

\end{abstract}

\textbf{Keywords}: Low-cost sensors, air pollution, calibration, deep learning, LSTM, feature engineering.  

%-------------------------------------------
% Paper Body
%-------------------------------------------
%--- Section ---%
%% main text
\section{Introduction\label{sec:introduction}}

Outdoor air pollution was responsible for an estimated 4.2 million premature deaths in 2019, with 89\% occurring in low- and middle-income countries \cite{who2024ambient}. Fine particulate matter (PM\textsubscript{2.5}), coarse particulate matter (PM\textsubscript{10}), and nitrogen dioxide (NO\textsubscript{2}) are among the most harmful urban pollutants, linked 
to respiratory illness, cardiovascular disease, and premature mortality \cite{who2024ambient, southerland2022global}. Levels of these pollutants change sharply by location, traffic, and time of year, illustrating the dynamic nature of urban air quality. Continuous, high spatio-temporal resolution monitoring is therefore essential to accurately estimate their behaviour and guide public health measures and policies \cite{karagulian2019review,ottosen2021perspectives}. However, the high cost and infrastructure requirements of reference-grade air quality monitoring stations have led to only sparse deployment \cite{liang2021calibrating}. In this context, low-cost sensors (LCS) have emerged as potentially viable alternatives, enabling real-time networks for urban air quality surveillance. Recent work has shown that real-time insights into urban air quality can be provided when LCSs are deployed alongside more sparse reference stations in a complementary fashion \cite{yin2025field}. Despite their appeal, LCS measurements are often compromised by environmental cross-sensitivity, temporal drift, and inter-sensor variability \cite{nalakurthi2024challenges,jourdan2021reliability}. These limitations introduce non-linearity and systematic bias, particularly under fluctuating ambient conditions such as temperature and humidity \cite{veiga2021blind}. Thus, reliable calibration is essential before these sensors can be integrated into decision-making processes or scientific studies. 

Calibration of low-cost air quality sensors has been approached using a range of statistical models. Regression methods remain the most widely applied, with univariate linear regression (ULR) serving as a benchmark and multivariate linear regression (MLR) improving accuracy by incorporating meteorological or co-pollutant predictors \cite{liang2021calibrating}. Traditional calibration of LCS has typically relied on co-location with reference-grade instruments and employed statistical methods. While effective in certain conditions, these approaches often fail to generalize across locations or over time due to their inability to capture the non-linearities and temporal dynamics inherent in sensor behaviour \cite{liang2021calibrating, si2020evaluation}. This emphasizes the need for improved and reliable techniques for LCS calibration.

Several studies have highlighted the importance of applying machine learning (ML) techniques to calibrate LCS outputs against reference instruments, reporting improved performance over traditional methods \cite{liang2021calibrating,yin2025field, si2020evaluation,park2021assessment}. Liang \cite{liang2021calibrating} provided a detailed description of various regression and ML approaches, reviewing state-of-the-art LCS calibration approaches, highlighting the shift from traditional statistical corrections to advanced data-driven models. Among these, supervised regression models such as Random Forest (RF), XGBoost, Support Vector Machines (SVM), and neural networks have been widely explored. For instance, Si et al. \cite{si2020evaluation} compared MLR, XGBoost, and feed-forward neural networks for PM$_{2.5}$ calibration, demonstrating that deep models consistently outperformed linear baselines, particularly under varying humidity conditions.  Nowack et al. \cite{nowack2021machine} demonstrated the successful calibration of co-located PM$_{10}$ and NO$_2$ sensors, using methods such as ridge regression, RF regression, and Gaussian process regression (GPR). Similarly, Yin et al. \cite{yin2025field} applied spatially-aware XGBoost models to improve generalization across urban sensor deployments. While Zimmerman et al. \cite{zimmerman2018machine} highlighted that machine learning methods, such as RF, can effectively mitigate drift and cross-sensitivity issues. 

Extending this line of work, Apostolopoulos et al.~\cite{apostolopoulos2023field} conducted an extensive field validation of low-cost NO$_2$ and O$_3$ sensors across urban deployment in Greece. They assessed machine learning approaches such as RF and LSTM models and found that incorporating meteorological context and carefully selecting input features significantly enhanced the calibration model’s ability to generalize across different environmental conditions. Spinelle et al.~\cite{spinelle2017field} demonstrated that supervised learning techniques performed better than traditional regression techniques. Han et al. \cite{han2023machine} described a multi-task methodology for air quality prediction, where temporal dynamics and data preprocessing are jointly leveraged to improve forecasting performance. Mahajan and Helbing \cite{mahajan2025dynamic} introduced an innovative trust-based calibration method that dynamically adapts model complexity based on sensor reliability. They employed trust-weighted consensus and wavelet-based features to calibrate PM$_{2.5}$ in real time, reducing mean absolute error by up to 68\% for low-quality sensors while maintaining computational efficiency. Bigi et al. \cite{bigi2018performance} and Bush et al. \cite{bush2021mlfield} have demonstrated that ensemble machine learning techniques, such as RF, are effective for calibrating low-cost air quality sensors. Bigi et al. \cite{bigi2018performance} evaluated three calibration strategies applied across a set of electrochemical sensors located in Switzerland. Their findings demonstrated significant reductions in mean absolute error when using ensemble methods, underscoring their portability and effectiveness across diverse environments. In particular, Bush et al. \cite{bush2021mlfield} proposed an RF regression pipeline to correct baseline drift and environmental interferences in NO\textsubscript{2}, PM\textsubscript{10}, and PM\textsubscript{2.5} measurements. The sensor drift and offset were handled by the adaptive iteratively reweighted penalised least squares (airPLS) techniques. However, despite its strengths, the RF method is inherently static and lacks memory of prior observations, which limits its capacity to generalize, model temporal drift, and long-term dynamics in sensor behaviour.

Feature engineering is another important part of a calibration model. Having well-chosen features is essential for any sensor calibration pipeline, as the model can only learn relationships that are represented in its inputs. Prior work shows that explicitly representing period and trend noticeably improves spatio–temporal modelling in urban settings \cite{zhang2017deep}. Modelling daily and weekly cycles enables learning recurrent pollutant dynamics, which are characteristic of urban air quality systems \cite{du2019deep}.

Deep learning methods that incorporate temporal structure, most notably Long Short-Term Memory (LSTM) networks, have recently gained traction for their ability to model sequence-dependent patterns in sensor drift and performance \cite{park2021assessment}. Recurrent neural networks (RNNs) often suffer from vanishing or exploding gradients, which limit their ability to capture long-range dependencies in sequential data. LSTM networks address this issue by using gated units that regulate how information is stored, updated, and discarded \cite{Goodfellow-et-al-2016}. Temporal sequences calibrated with an LSTM network are well-suited to learning delayed effects and persistence in sensor signals \cite{hochreiter1997long}. These models view calibration not merely as static regression, but as a temporal inference task that evolves in response to environmental variability and site-specific dynamics.
%\cite{bush2021mlfield, mahajan2025dynamic}.

For example, Park et al. \cite{park2021assessment} developed a hybrid LSTM model for PM$_{2.5}$ calibration, combining meteorological inputs such as temperature and relative humidity, and demonstrated superior capture of pollutant dynamics compared to feed-forward models. Veiga et al. \cite{veiga2021blind} extended this by applying LSTM (alongside CNNs) in blind wireless sensor network calibration, showing marked error reductions without reference data. Beyond particulate matter, other studies highlight LSTM’s efficacy in calibrating gaseous pollutants. Han et al. \cite{han2021calibrations} demonstrated that LSTM significantly improved NO$_2$ accuracy, achieving $R^2$ greater than 0.7 for the test set, surpassing RF and linear models in field conditions. Ryu and Park \cite{ryu2022band} introduced a band-sensitive LSTM that dynamically reweights errors in critical concentration ranges, reducing RMSE by 12\% and improving tail accuracy. These examples underscore the flexibility of LSTM architectures for diverse pollutants and calibration scenarios.

Meta-learning strategies are also garnering attention. Yadav et al. \cite{yadav2022few} proposed a few-shot calibration setup using Model-Agnostic Meta-Learning (MAML), enabling rapid calibration of new sensors with minimal co-location data. This paradigm addresses one of the central bottlenecks of ML-based calibration, which is the dependency on extensive labelled training data. Furthermore, Di Antonio et al. \cite{di2018developing} focused on correcting PM measurements distorted by relative humidity. By incorporating a physical correction based on particle hygroscopicity, they drastically reduced PM$_{2.5}$ overestimation. Alternatively, Patel et al.~\cite{patel2024towards} proposed a hygroscopic-growth calibration for low-cost nephelometric PM$_{2.5}$ sensors, modelling the seasonal evolution of kappa($\kappa$) to correct RH-dependent biases.

Additional studies further underscore the heterogeneity in LCS performance. Atfeh et al. \cite{atfeh2025performance} performed a comparative assessment of PM$_{2.5}$ sensors under real-world conditions in Central Europe, noting that while many devices correlated well with reference monitors, bias and precision issues remained. They recommend hybrid techniques combining empirical correction and ensemble ML to improve accuracy. Other promising directions include AI-IoT integration. Rakib et al.~\cite{rakib2022iot} developed an IoT-based framework for air quality monitoring and 24-hour PM$_{2.5}$ forecasting using an ARIMA model, demonstrating the potential scalability of cloud-connected sensor networks in real-time applications.

The above studies demonstrate significant progress in sensor calibration but also reveal persistent challenges. The proposed work builds and improves upon the calibration framework developed by Bush et al. \cite{bush2021mlfield}. The calibration framework proposed by Bush et al. \cite{bush2021mlfield} relies on RF regression, which treats each observation independently and therefore cannot explicitly capture temporal dependencies, sensor drift, or delayed environmental effects that influence low-cost sensor behaviour. These affect generalization to unseen data and have motivated the present work, which introduces a temporal deep learning calibration framework targeting PM\textsubscript{2.5}, PM\textsubscript{10}, and NO\textsubscript{2} measurements. Several recent studies have shown that sequence-based deep learning models such as Long Short-Term Memory (LSTM) networks can improve calibration accuracy over traditional ML methods by modelling temporal dynamics in environmental sensor data \cite{ park2021assessment, han2021calibrations, li2016deep}. Moving from point-wise prediction to sequence modelling allows the calibration framework to incorporate temporal context, enabling the model to learn persistence, delayed environmental effects, and gradual sensor drift that cannot be captured when each observation is treated independently. A hypertuned LSTM model has been developed and trained on co-located datasets, integrating lagged pollutant values and meteorological parameters. By incorporating rolling temporal windows and lagged environmental and pollutant features, the proposed method captures temporal dynamics in sensor responses and improves generalization performance on unseen datasets.

The remainder of this paper is structured into four sections. Section~2 presents the methodology, covering data preparation, feature engineering, model architecture, and training procedures. Section~3 outlines the test case, and Section~4 reports the results, with performance metrics, comparisons against baseline models, and detailed error analysis. Finally, Section~5 concludes the study by summarizing key findings and directions for future research.

%--- Section ---%
\section{Methodology}\label{sec2}

This section describes the steps followed in developing and evaluating the proposed calibration framework. The reliability of a calibration model comes not just from the choice of the model or architecture, but from how the supporting elements are designed. The design must ensure that sensor data are represented in a way that preserves their temporal structure and captures the environmental conditions under which they operate. Incorporating physically meaningful variables such as temperature, humidity, or flow dynamics allows the model to learn relationships grounded in sensor behaviour rather than relying on purely statistical correlations. Equally important is the selection of network parameters. Proper tuning of model parameters like window size, learning rate, and batch size ensures that the model captures long-term dependencies while remaining stable and generalizable.  

With these considerations, a calibration pipeline has been developed that treats LCS signals as temporally evolving processes, shaped by both environmental conditions and intrinsic sensor behaviour. The methodology has been summarized in Fig.~\ref{fig:meth}. This consists of the following steps:
\begin{itemize}
    \item \textbf{(a) Raw Sensor Data Input:} Time-stamped measurements from low-cost devices (PM$_{2.5}$, PM$_{10}$, NO$_2$) are combined with meteorological variables and co-located reference observations to form the initial dataset.
    
    \item \textbf{(b) Feature Engineering \& Preprocessing:} Lagged and harmonic signals, domain-specific interactions, and traffic-related effects such as rush-hour are incorporated in the model.
    
    \item \textbf{(c) LSTM Calibration Model:} A hypertuned LSTM network processes fixed-length sequences of input data, capturing sensor drift and delayed pollutant responses. 
    
    \item \textbf{(d) Calibrated Output:} Error metrics are then employed to evaluate the calibrated sensor output.
\end{itemize}

\begin{figure}[h!]
  \centering
  \includegraphics[width=\textwidth]{meth.jpg}
  \caption{Overview of the calibration methodology: (a) raw sensor data input, (b) feature engineering and preprocessing, (c) LSTM calibration model, and (d) calibrated output evaluation.}
  \label{fig:meth}
\end{figure}

\subsection{Data Structure}

The dataset analysed in this study is derived from the OxAria project \cite{Bush2022OxAria}, a large-scale deployment of low-cost air quality sensors across the city of Oxford. The measurements include particulate matter (PM$_{2.5}$ and PM$_{10}$) and nitrogen dioxide (NO$_2$), alongside meteorological parameters such as ambient temperature and relative humidity. Depending on the pollutant channel, additional auxiliary variables are recorded. For electrochemical NO$_2$ sensing, these include working and auxiliary electrode voltages, which provide information on baseline stability and environmental sensitivity. For particulate matter channels, flow-related variables such as sample flow rate and time-of-flight metrics are included, characterising transport through the optical chamber.

An important aspect of the dataset used in this work is the baseline correction applied before analysis. As mentioned in the previous section, low-cost sensors frequently exhibit signal drift, offset variability, and spurious fluctuations due to environmental influences such as humidity or temperature. To address these issues, Bush et al. ~\cite{bush2021mlfield} implemented a four-stage preprocessing routine. The procedure begins with empirical filtering to remove physically implausible values and transient artefacts. This is followed by the application of airPLS, which captures and subtracts slow-varying offsets without suppressing short-term pollution dynamics. A compensation step is then introduced to prevent small over-corrections and residual anomalies, such as negative concentrations introduced by the correction. Baseline correction removes long-term drift and offsets that would otherwise distort the true pollution signal.

This pipeline produces corrected sensor time series that are substantially more stable and reliable, while retaining the variability necessary for calibration against reference monitors. Bush et al. ~\cite{bush2021mlfield} reported that after correction, mean absolute errors of the Praxis Urban sensors are reduced, greatly improving the quality of the input data for machine learning applications. In the present study, this corrected dataset was used directly as the foundation for the model development. By relying on this preprocessed data, the analysis is able to concentrate on the design and evaluation of the temporal calibration rather than the intricacies of baseline adjustment.

Table~\ref{tab:featuretable} summarizes the input variables used in model training by pollutant type.

\begin{table}[h!]
\centering
\begin{tabular}{lccc}
\toprule
\textbf{Feature (Variable)} & \textbf{NO$_2$} & \textbf{PM$_{10}$} & \textbf{PM$_{2.5}$} \\
\midrule
Sensed concentration / mass (conc)   & \checkmark & \checkmark & \checkmark \\
Working electrode voltage (wev)      & \checkmark & --         & --         \\
Auxiliary electrode voltage (aev)    & \checkmark & --         & --         \\
Sample flow rate (sfr)               & --         & \checkmark & \checkmark \\
Sample time of flight (mtf)          & --         & \checkmark & \checkmark \\
Temperature (tmp)                    & \checkmark & \checkmark & \checkmark \\
Relative humidity (hmd)              & \checkmark & \checkmark & \checkmark \\
\bottomrule
\end{tabular}
\caption{Sensor variables used for model training and prediction by pollutant type (adapted from Bush et al.~\cite{bush2021mlfield}).}
\label{tab:featuretable}
\end{table}

This pollutant-specific feature matrix forms the foundation for the DL model trained in this study, ensuring consistency with the RF benchmarks used in prior work. In the LSTM-based model, additional temporal features and lag structures are incorporated during preprocessing, as described in the following sections.

\subsection{Feature Engineering and Model Design} 

The engineered features focus on constructing a comprehensive input representation that reflects sensor behaviour, environmental drivers, and temporal structure. The model design integrates these features into a sequence-based learning pipeline, which includes data partitioning, normalization, sequence generation, and hyperparameter tuning. 

\subsubsection{Feature Engineering}
To capture temporal dynamics and environmental influences, a rich set of features was engineered from the baseline-corrected LCS dataset. These include both direct sensor observations and derived variables (Tab. \ref{tab:engineered_features}). Here, the final input matrix $\boldsymbol{X} \in \mathbb{R}^{\boldsymbol{T} \times \boldsymbol{F}}$ consists of $\boldsymbol{F}$ features observed over $\boldsymbol{T}$ time steps, grouped into the following categories:

\begin{itemize}
    \item \textbf{Raw Sensor Signals:}  
    Direct outputs from the Praxis units, including particulate concentrations, sample flow rate, time-of-flight metrics, ambient temperature, relative humidity, etc. These signals carry pollutant information but also embed sensor drift and environmental cross-sensitivities that require calibration.
    
\item \textbf{Temporal Gradient Features:}  
    To capture rapid fluctuations, first- and second-order percentage changes are computed \cite{bush2021mlfield}:  
    \begin{equation}
     pc^{15}_x(t) = \frac{x(t) - x(t-1)}{x(t-1)}, \qquad
     pc^{30}_x(t) = \frac{x(t) - x(t-2)}{x(t-2)}
     \label{eq:temporal_gradients}
    \end{equation}
    where $x(t)$ denotes the value of a signal at time $t$. These emphasize short-term variability that raw concentrations may not fully capture.

    \item \textbf{Temporal and Cyclical Encodings:}  
    Time-of-day, day-of-week, and rush-hour indicators are included to capture diurnal and weekly cycles. To maintain continuity across cycle boundaries, harmonic features are encoded as:
    \begin{equation}
\begin{split}
\mathrm{hour}^{\sin}(t) &= \sin\left(\frac{2\pi h(t)}{24}\right), \quad
\mathrm{hour}^{\cos}(t) = \cos\left(\frac{2\pi h(t)}{24}\right) \\
\mathrm{day}^{\sin}(t) &= \sin\left(\frac{2\pi d(t)}{7}\right), \quad
\mathrm{day}^{\cos}(t) = \cos\left(\frac{2\pi d(t)}{7}\right)
\end{split}
\label{eq:cyclical_features}
\end{equation}
    with $h(t)$ and $d(t)$ denoting hour and weekday, respectively. Monthly and seasonal cycles are similarly encoded. These features help the model recognize recurring temporal patterns in air pollution and improve consistency during repetitive events such as rush hours or seasonal shifts.

    \item \textbf{Interaction Features:}  
    Interaction is computed between various features available in the dataset. For example, the ratio \( mtf/(tmp+\epsilon) \) and the product \( sfr \cdot hmd \), where $\epsilon$ is a small constant to avoid division by zero. These interactions reflect the combined influence of meteorology and particle optics on sensor performance \cite{castell2017can}.
    
    \item \textbf{Target Lags:}  
    Recent lagged values of the pollutant under calibration and their percentage changes are included to provide autoregressive context. This allows the model to account for persistence and short-term correlation in pollutant levels.
\end{itemize}

\begin{table}[h!]
\centering
\begin{tabular}{ll}
\toprule
\textbf{Category} & \textbf{Examples} \\
\midrule
Raw Sensor Values & conc, sfr, mtf, tmp, hmd \\
Lagged Changes & pc$^{15}$tmp, pc$^{30}$mtf, pc$^{30}$hmd etc. \\
Time Features & hour, day, rush hour \\
Cyclical Encodings & hour$^{\sin}$, day$^{\cos}$, month$^{\sin}$, season$^{\cos}$ etc. \\
Sensor Interactions & mtf/tmp, sfr$\cdot$hmd etc. \\

\bottomrule
\end{tabular}
\caption{Summary of raw and engineered features included in the calibration pipeline.}
\label{tab:engineered_features}
\end{table}

\subsubsection{Sequence Construction and Hyperparameter Tuning}

Air pollutant dynamics and LCS readings depend not only on instantaneous conditions but also on recent history. To model this dependency, a supervised sequence-to-one learning strategy is adopted using a fixed-length rolling window technique. Let $\boldsymbol{x}_t \in \mathbb{R}^{F}$ denote the engineered feature vector at time $t$ (all inputs except the target channel), and let $\boldsymbol{y}_t \in \mathbb{R}$ be the co-located reference value used as the label. The dataset is then converted into overlapping input–output pairs as
\begin{equation}\label{eq:RoW}
\boldsymbol{X}_i = \big[\boldsymbol{x}_i,\, \boldsymbol{x}_{i+1},\, \dots,\, \boldsymbol{x}_{i+W-1}\big] \in \mathbb{R}^{W \times F},
\qquad
\boldsymbol{y}_i = \boldsymbol{y}_{i+W},
\qquad i = 1, 2, \dots, N,
\end{equation}
where $W$ is the sequence length (window size), $T$ the total number of time steps, and $N = T-W$ the total number of training pairs. This rolling-window setup preserves recent temporal context while also providing a fixed input size \cite{amor2016recursive,li2014rolling}. Each window of $W$ past measurements is mapped to the next-step target, and the window then slides forward by one step to form the next training pair. It ensures that each time step contributes both as part of an input sequence and, one step later, as a label.

%In code, the input matrix is built by selecting all features except the target signal, and the rolling-window generator produces the tensors $\boldsymbol{X}_{\text{seq}}\in\mathbb{R}^{\boldsymbol{N}\times \boldsymbol{W}\times \boldsymbol{F}}$ and $\boldsymbol{y}_{\text{seq}}\in\mathbb{R}^{\boldsymbol{N}}$. A single-step horizon ($\boldsymbol{t}\!\to\! \boldsymbol{t}{+}1$) is used, which is natural for calibration and avoids compounding forecast errors while still letting the model learn delayed responses.

%\begin{figure}[h!]
%  \includegraphics[width=\textwidth]{RoW.jpg}
%  \caption{Rolling-window sequence construction: each length-$\boldsymbol{W}$ input sequence (green) maps to the next-step target (blue).}
%  \label{fig:RW}
%\end{figure}

The effectiveness of sequence models such as LSTMs depends strongly on their hyperparameters, which control both the structure of the network and the dynamics of training. To identify a stable and accurate configuration, a grid search was carried out over three key parameters, ${W}$, learning rate (${\eta}$), and batch size.

%\begin{itemize}
    %\item \textbf{Window Size} ($\boldsymbol{W}$): Window size is the number of past time steps included in each input sequence, defining the temporal context available to the model.
    %\item \textbf{Learning Rate} ($\boldsymbol{\eta}$): Learning rate determines the step size used by the Adam optimizer to update network weights, which influences both convergence speed and stability.
    %\item \textbf{Batch Size}: Batch size describes the number of training samples processed before updating model parameters, balancing computational efficiency with gradient accuracy.
%\end{itemize}%

Grid search systematically evaluates all possible combinations of these parameters within predefined ranges and selects the configuration that yields the best validation performance \cite{bergstra2012random}.

\subsection{LSTM-Based Calibration Model}

In this application, the input is a normalized (Eq.~\ref{eq:zscore_normalization}), fixed-length rolling window (Eq.~\ref{eq:RoW}) of engineered features (Eqs.~\ref{eq:temporal_gradients} and \ref{eq:cyclical_features}), and the target is the reference observation. The network is implemented as a combination of multiple layers. A single LSTM block, followed by a dense layer with Leaky-ReLU ($\alpha=0.01$), a Dropout layer, and a final linear unit (Tab. \ref{tab:lstm-arch}) for the calibrated concentration. The model is trained with Adam using the mean absolute error objective with early stopping on validation loss. 
%Formally, for a window length $W$ and $F$ features per step,

%\begin{equation}
%\label{eq:lstm_mapping}
%\boldsymbol{X}_{t-W+1:t} \in \mathbb{R}^{W \times F}
%\;\longrightarrow\;
%\hat{y}_{t+1} = f_{\boldsymbol{\theta}}\!\left(\boldsymbol{X}_{t-W+1:t}\right),
%\end{equation}

%where $f_{\boldsymbol{\theta}}$ is the LSTM model.

\begin{equation}
    \text{MAE} = \frac{1}{n} \sum_{i=1}^{n} \left| \mathbf{y}_i - \mathbf{\hat{y}}_i \right|
    \label{eq:mae}
\end{equation}

where \( \mathbf{y}_i \) represents the true values, \( \mathbf{\hat{y}}_i \) denotes the predicted values, and \( n \) is the total number of steps.

\begin{table}[h!]
\centering

\begin{tabular}{llll}
\toprule
\textbf{Layer} & \textbf{Layer Details} & \textbf{Units / Params} & \textbf{Activation / Output Dim.} \\
\midrule
0 & Input & -- & $(W,F)$ \\
1 & LSTM & 128 &  $ \in\mathbb{R}^{128}$ \\
2 & Dense & 64 & Leaky-ReLU ($\alpha=0.01$); $ \in\mathbb{R}^{64}$ \\
3 & Dropout & rate $=0.3$ & $ \in\mathbb{R}^{64}$ \\
4 & Dense (output) & -- & Linear; $ \hat{y}_{t+1}\in\mathbb{R}$ \\
\bottomrule
\end{tabular}
\caption{Layer specification for the LSTM calibration network (matching the implemented model).}
\label{tab:lstm-arch}
\end{table}

With such rich engineered inputs, the network has sufficient capacity to overfit these idiosyncrasies. L2 regularization ($\lambda=10^{-4}$) discourages large weights, which helps smooth the function and makes it generalize more reliably. Dropout prevents over-reliance on individual neurons, effectively acting like an ensemble to reduce variance \cite{srivastava2014dropout}. Together with early stopping, these controls produce a stable calibrator that maintains accuracy when applied to hold-out and unseen periods.

\subsection{Evaluation Metrics}

The performance of the calibration strategy is assessed using three widely adopted statistical metrics: the coefficient of determination ($R^2$), the Mean Absolute Error (MAE), and the Root Mean Squared Error (RMSE). These metrics capture different aspects of calibration accuracy.

\begin{itemize}
    \item \textbf{$R^2$ (Coefficient of Determination):}  
    $R^2$ values quantify the proportion of variance in the reference observations explained by the predictions:  
    \begin{equation}\label{eq:R21}
    R^2 = 1 - \frac{\sum_{i=1}^{N}(y_i - \hat{y}_i)^2}{\sum_{i=1}^{N}(y_i - \bar{y})^2}
    \end{equation}
    where $y_i$ are reference values, $\hat{y}_i$ are the calibrated values, and $\bar{y}$ is the mean of the reference set. A value close to 1 indicates strong explanatory power \cite{article}.
    
    \item \textbf{MAE (Mean Absolute Error):}  
    MAE represents the average absolute deviation between predicted and observed values (Eq. \ref{eq:mae}). MAE provides a straightforward measure of typical prediction error and is less influenced by extreme outliers \cite{article}.
    
    \item \textbf{RMSE (Root Mean Squared Error):}  
    RMSE emphasizes the larger deviations by squaring residuals before averaging:  
    \begin{equation}\label{eq:RMSE}
    \mathrm{RMSE} = \sqrt{\frac{1}{N} \sum_{i=1}^{N} \left( y_i - \hat{y}_i \right)^2}
    \end{equation}
    RMSE is particularly relevant for air quality studies, as it penalizes missed peaks in pollutant concentrations \cite{article}.
    
\end{itemize}

Together, these three metrics provide a balanced evaluation of calibration performance. $R^2$ (Eq. \ref{eq:R21}) captures explanatory power, MAE (Eq. \ref{eq:mae})  and RMSE (Eq. \ref{eq:RMSE}) quantify error magnitudes under different sensitivities.

\section{Test Case}

The OxAria LCS network deployed 16 low-cost sensor units across diverse urban micro-environments, with the goal of enabling high-resolution spatio-temporal air quality monitoring.  This network provides coverage of key urban pollutants, complementing the sparse but highly accurate reference stations maintained under the UK’s Automatic Urban and Rural Network (AURN) programme. Each LCS unit is based on the Praxis Urban sensing platform from South Coast Science Ltd, integrating an Alphasense NO$_2$-A43F electrochemical sensor and an Alphasense N3 optical particle counter (OPC). The present study focuses on a single device co-located with the reference-grade AURN reference site at Oxford St.~Ebbe’s (UKA00518). Reference data span different periods for each pollutant, May 2020 to September 2021 for PM${10}$ and PM${2.5}$, and May 2020 to May 2021 for NO$_2$. The co-location ensures that both the LCS device and the reference station are exposed to identical atmospheric conditions, enabling reliable supervised calibration. The setup has been summarized in Tab. \ref{tab:oxaria-setup}.

\begin{figure}[h!]
    \centering
    
    \begin{subfigure}{\textwidth}
        \centering
        \includegraphics[width=\linewidth]{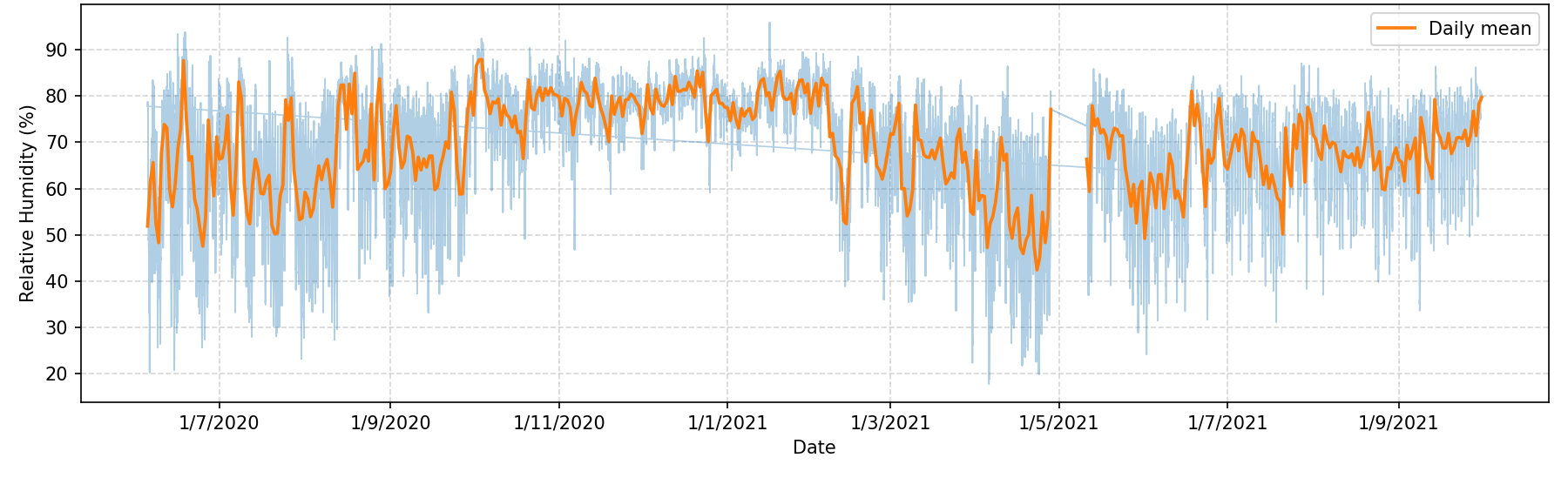}
        \caption{Relative humidity}
        \label{fig:hmd}
    \end{subfigure}
    
    \vspace{0.5em}
    
    \begin{subfigure}{\textwidth}
        \centering
        \includegraphics[width=\linewidth]{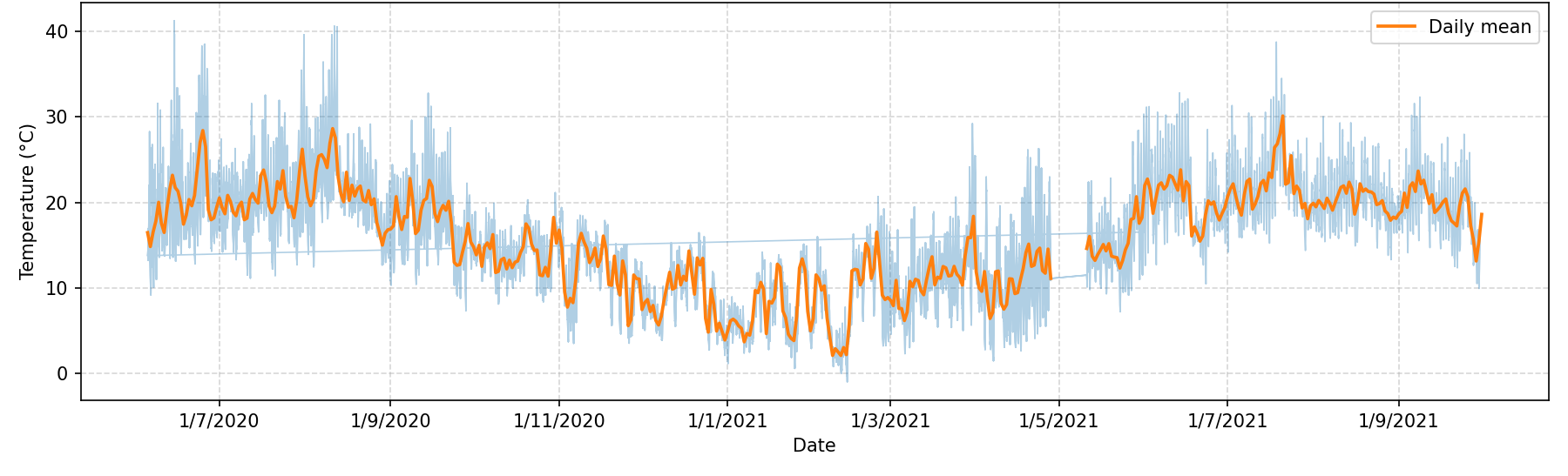}
        \caption{Temperature}
        \label{fig:tmp}
    \end{subfigure}
    
    \caption{Time series of meteorological variables: (a) Relative humidity and (b) temperature.}
    \label{fig:meteo}
\end{figure}

\begin{figure}[h!]
    \centering
    
    \begin{subfigure}{\textwidth}
        \centering
        \includegraphics[width=\textwidth]{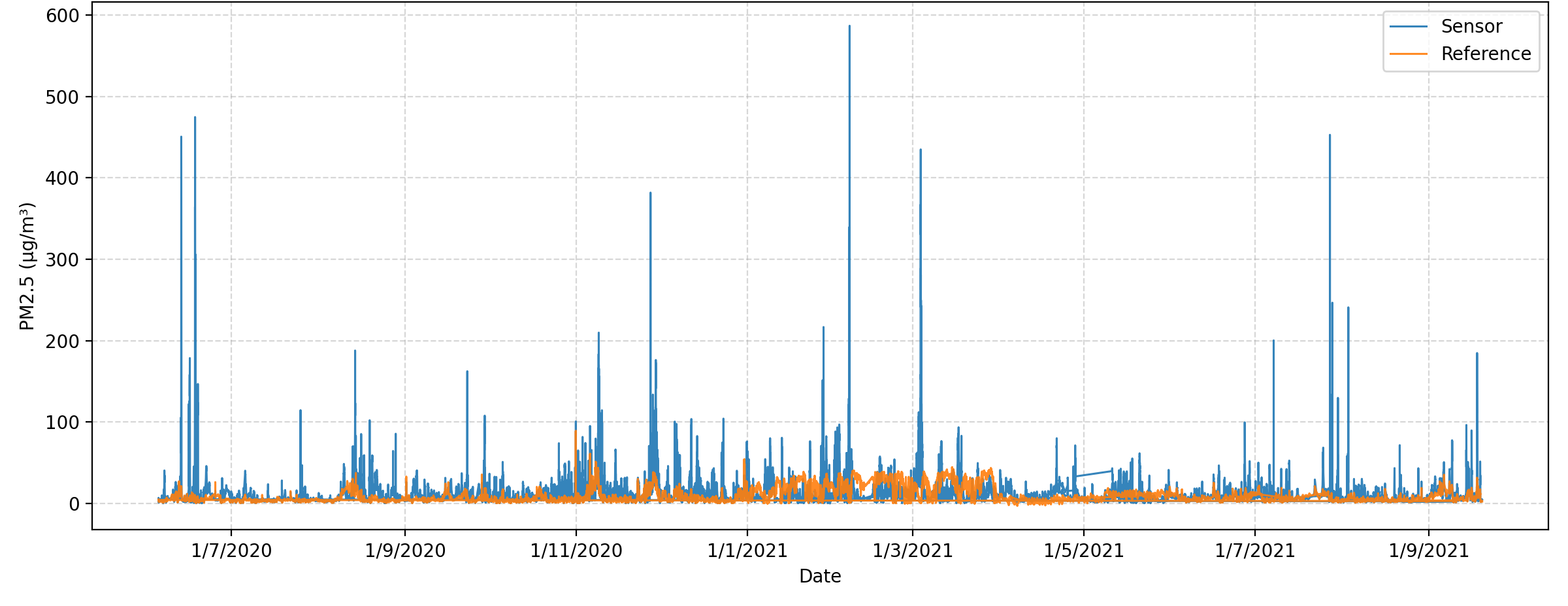}
        \caption{PM$_{2.5}$ sensor values against AURN reference at Oxford St.~Ebbe’s.}
        \label{fig:pm25}
    \end{subfigure}

    \begin{subfigure}{\textwidth}
        \centering
        \includegraphics[width=\textwidth]{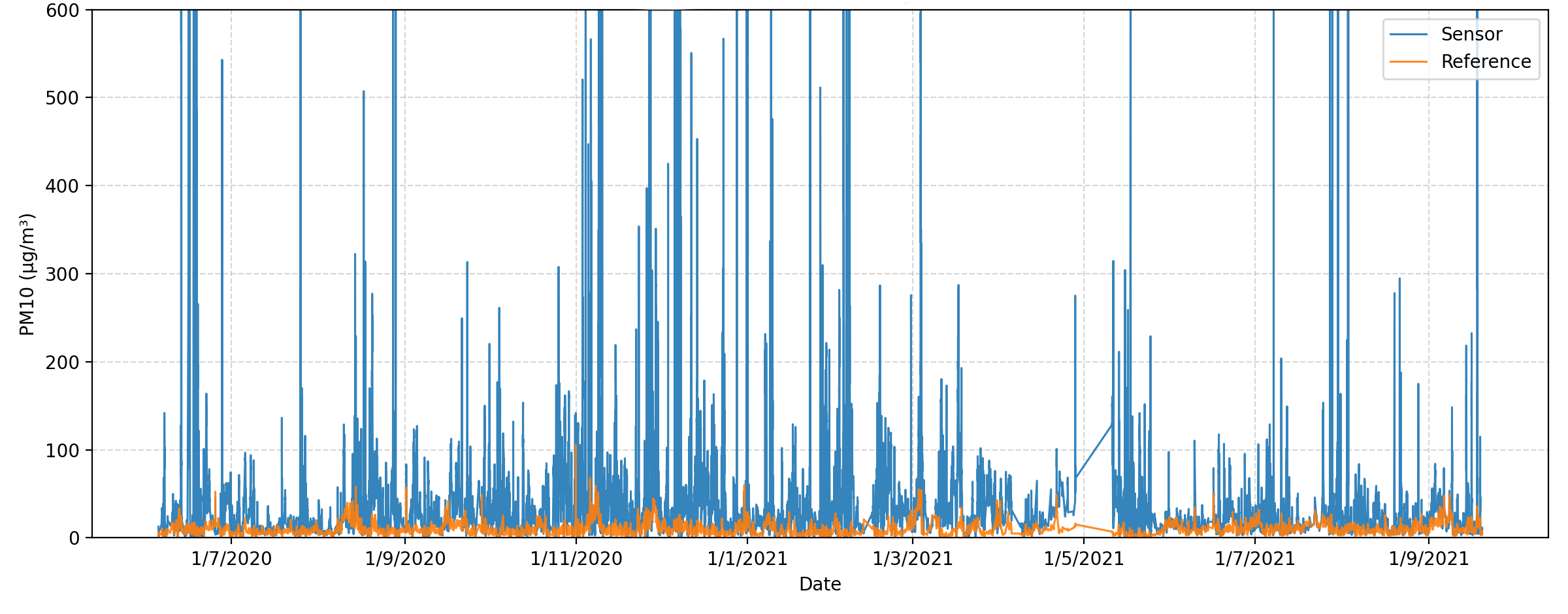}
        \caption{PM$_{10}$ sensor values against AURN reference at Oxford St.~Ebbe’s.}
        \label{fig:pm10}
    \end{subfigure}

    \begin{subfigure}{\textwidth}
        \centering
        \includegraphics[width=\textwidth]{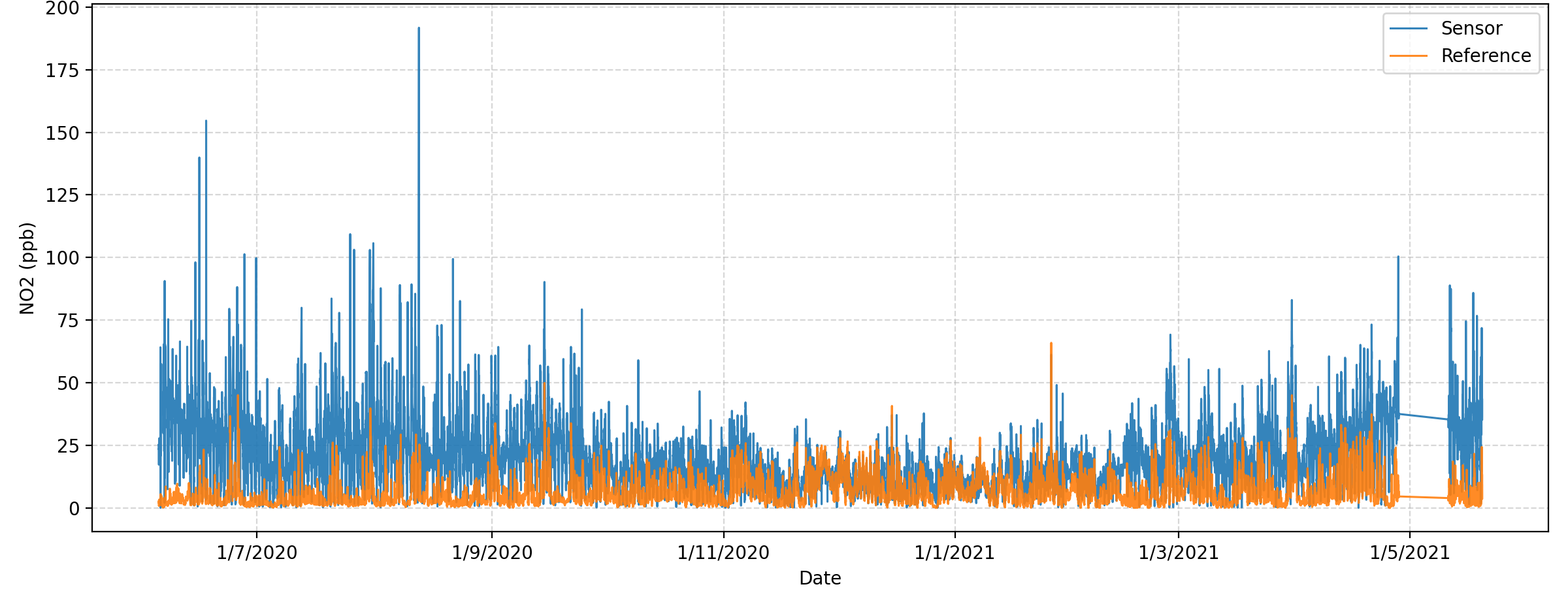}
        \caption{NO$_2$ sensor values against AURN reference at Oxford St.~Ebbe’s.}
        \label{fig:no2}
    \end{subfigure}
    
    \caption{Comparison of low-cost sensor measurements with AURN reference data at Oxford St.~Ebbe’s for (a) PM$_{2.5}$, (b) PM$_{10}$, and (c) NO$_2$.}
    \label{fig:baseline_comparison}
\end{figure}

Figure~\ref{fig:meteo} illustrates the temporal evolution of temperature and relative humidity during the observation period, demonstrating the variability that low-cost sensors are exposed to in the field. Figure ~\ref{fig:baseline_comparison} shows the time series for PM$_{2.5}$, PM$_{10}$, and NO$_2$, comparing AURN reference values with the sensor signals.

\begin{table}[h!]
\centering

\begin{tabular}{ll}
\toprule
\textbf{Parameter} & \textbf{Value} \\
\midrule
Sensor platform       & Praxis Urban (South Coast Science Ltd.) \\
Pollutants measured   & NO$_2$, PM$_{10}$, PM$_{2.5}$ \\
Sensor types          & Electrochemical (NO$_2$), OPC (PM) \\
Reference station     & Oxford St. Ebbe’s (AURN, UKA00518) \\
Reference instruments & Teledyne T200 (NO$_2$), Palas FIDAS 200 (PM) \\
Data resolution       & 15-minute averages \\
\bottomrule
\end{tabular}
\caption{Summary of the OxAria test case setup used for calibration experiments.}
\label{tab:oxaria-setup}
\end{table}

\section{Results}

This section presents the results of the proposed LSTM-based calibration approach for the low-cost air quality sensors. First, the scatter plots are presented, followed by the results of the evaluation metrics and the comparison of temporal predictions. The tuned hyperparameters used in the final model are summarized in~\ref{app:A2}.

\subsection{Pollutant-wise Performance Evaluation}

Since the behaviour and error characteristics of LCS vary across pollutants, the calibrated performance has been analyzed separately for PM$_{2.5}$, PM$_{10}$, and NO$_2$. This pollutant-wise breakdown highlights how the LSTM setup adapts to distinct sensing challenges. For each case, the model scatter plots, calibration accuracy against reference instruments, and a comparison of error statistics are presented in the sections below.

\subsubsection{Scatter Plots}

\begin{figure}[h!]
\centering

\begin{subfigure}{0.43\textwidth}
\centering
\includegraphics[width=\linewidth]{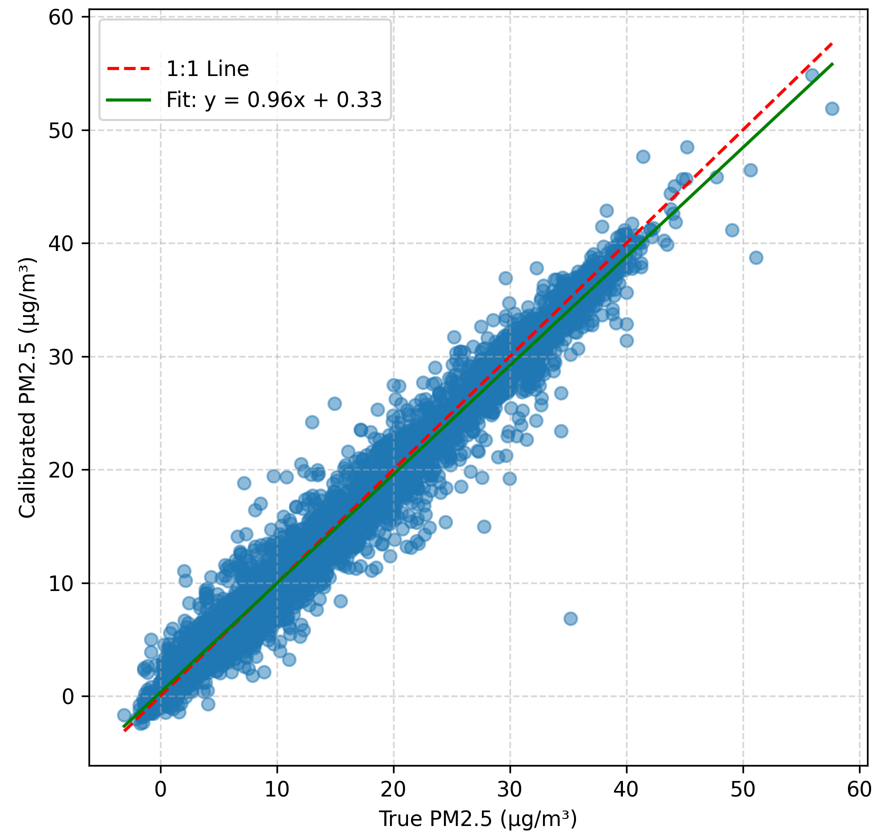}
\caption{PM$_{2.5}$ (Validation)}
\end{subfigure}
\hfill
\begin{subfigure}{0.43\textwidth}
\centering
\includegraphics[width=\linewidth]{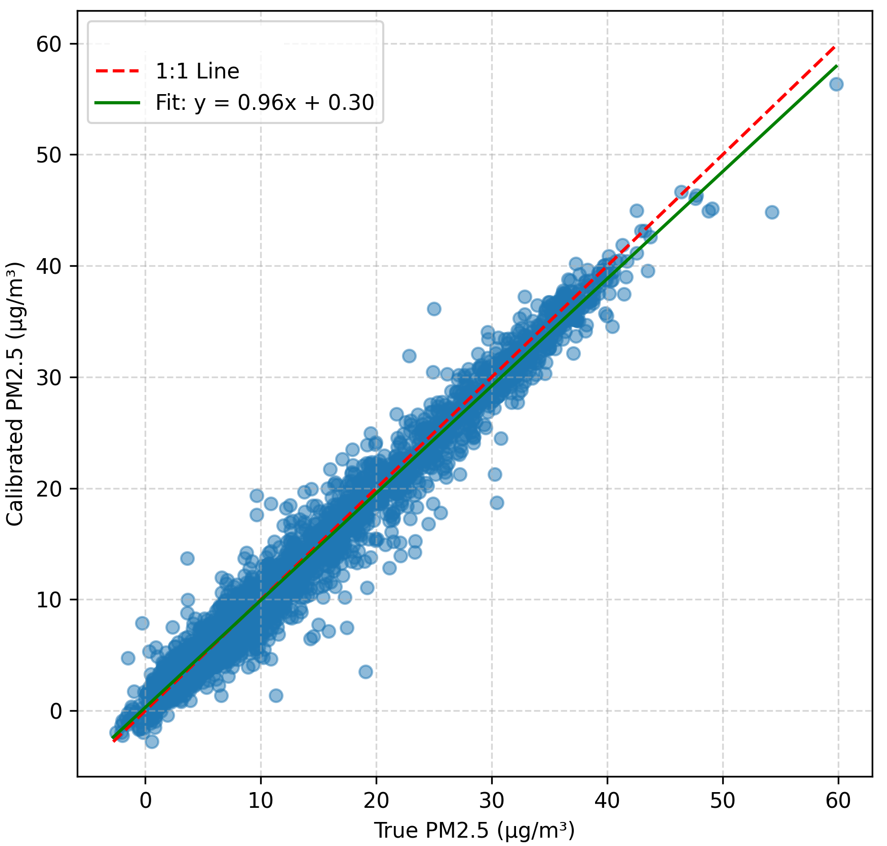}
\caption{PM$_{2.5}$ (Test)}
\end{subfigure}

\begin{subfigure}{0.43\textwidth}
\centering
\includegraphics[width=\linewidth]{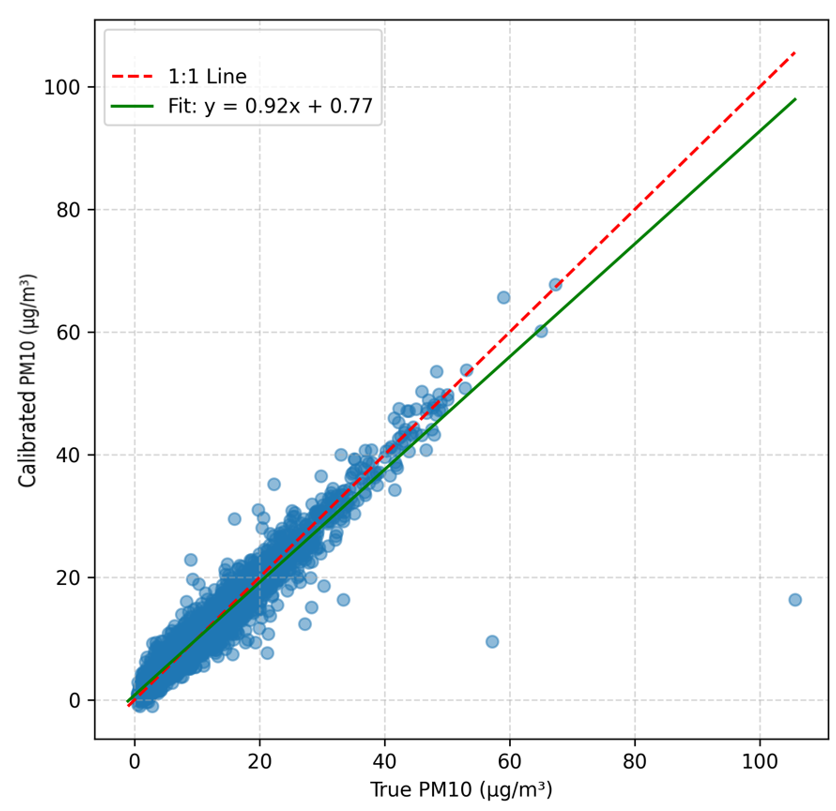}
\caption{PM$_{10}$ (Validation)}
\end{subfigure}
\hfill
\begin{subfigure}{0.43\textwidth}
\centering
\includegraphics[width=\linewidth]{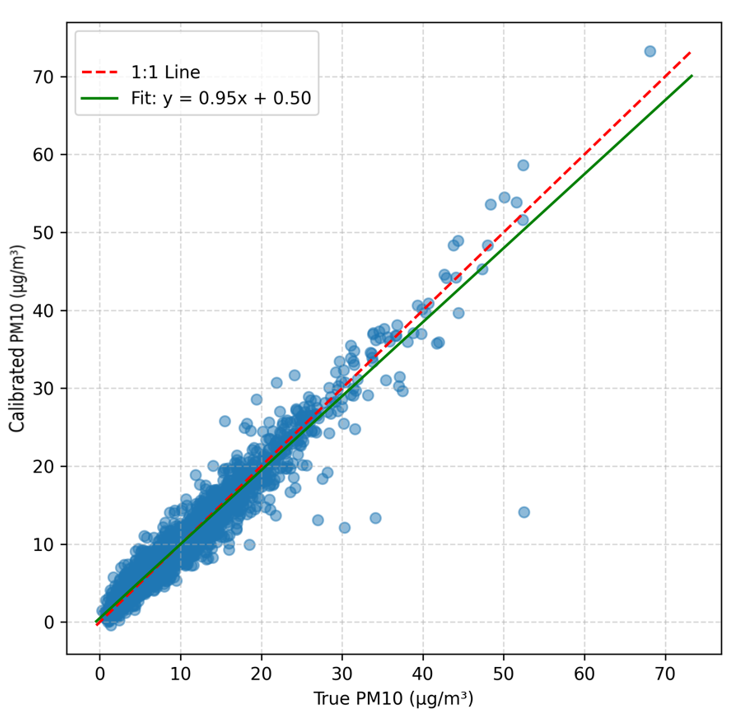}
\caption{PM$_{10}$ (Test)}
\end{subfigure}

\begin{subfigure}{0.43\textwidth}
\centering
\includegraphics[width=\linewidth]{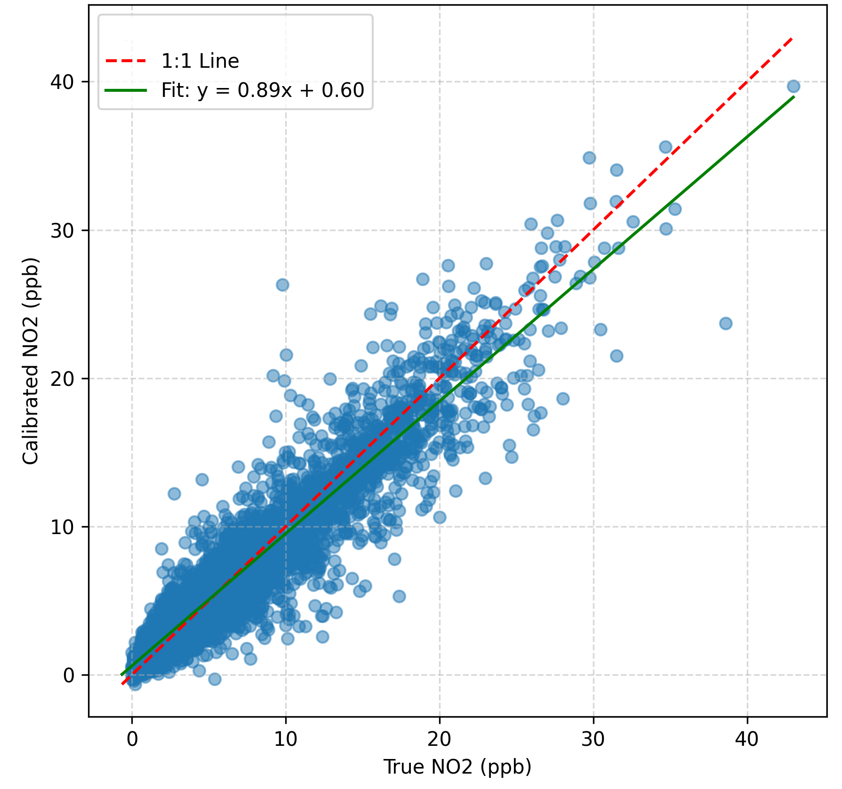}
\caption{NO$_2$ (Validation)}
\end{subfigure}
\hfill
\begin{subfigure}{0.43\textwidth}
\centering
\includegraphics[width=\linewidth]{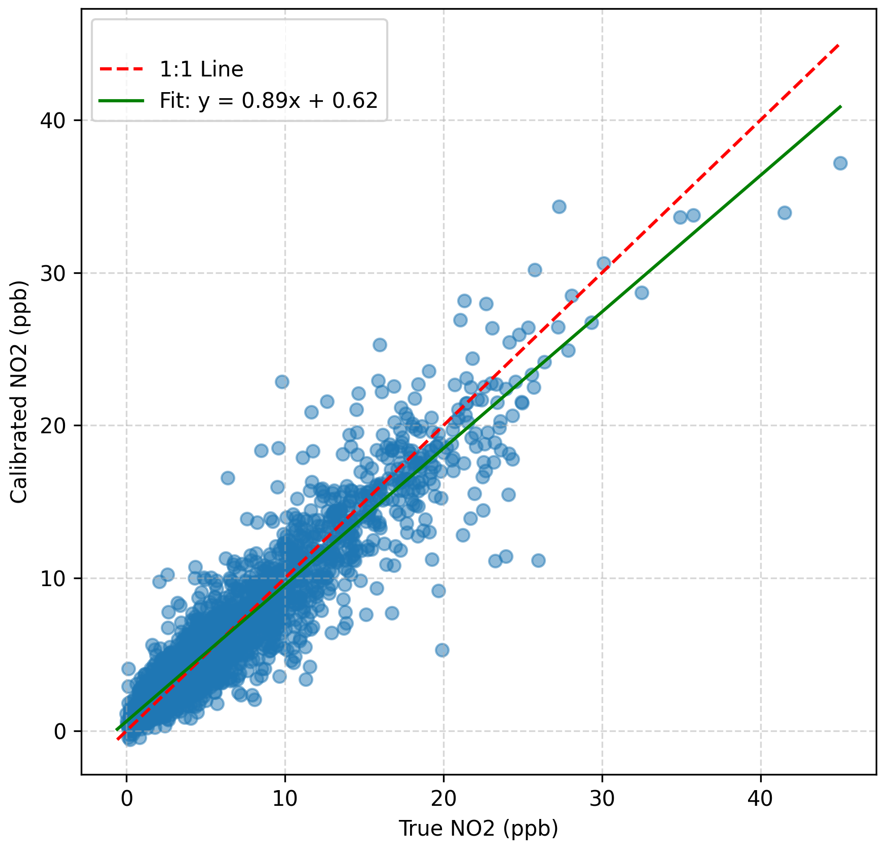}
\caption{NO$_2$ (Test)}
\end{subfigure}

\caption{Calibrated vs. reference concentrations across validation and test datasets for PM$_{2.5}$, PM$_{10}$, and NO$_2$. }
\label{fig:calibration_scatter}
\end{figure}

%Observations of fine particulate matter (PM$_{2.5}$) by OPC based LCS are highly sensitive to meteorological conditions and sensor drift, making it a central pollutant for calibration testing. 
Scatter plots for PM\textsubscript{2.5}, PM\textsubscript{10}, and NO\textsubscript{2} (Fig. \ref{fig:calibration_scatter}) show that the calibrated predictions follow the 1:1 line closely across all data splits. The red dashed line indicates the 1:1 reference (perfect fit), and the green line represents a linear regression fit \cite{harris2020array} between the calibrated and reference values. For each pollutant, the points form a tight cluster around the 1:1 line, and the fitted slopes remain close to one, indicating very little bias. This behaviour is consistent across the sets, suggesting that the model maintains its accuracy. Taken together, the results show stable calibration performance for all three pollutants, with strong alignment between the reference and the fitted regression across splits.

Table~\ref{tab:calibration_metrics} summarises the error metrics for all three pollutants. PM$_{2.5}$ shows the strongest performance, with $R^2$ values above 0.97 and low MAE and RMSE across all splits. The calibration performs well for PM$_{10}$, though generalization is slightly weaker than PM$_{2.5}$. The model still captures most of the variance, and the error levels between validation and test sets remain close, which indicates stable behaviour. For NO$_2$, the $R^2$ values stay above 0.88, and both MAE and RMSE remain within a narrow range, showing that the model tracks the main concentration patterns despite the higher variability typical of NO$_2$ sensors.

\begin{table}[h!]
\centering
\begin{tabular}{l l c c c}
\toprule
\textbf{Pollutant} & \textbf{Split} & $R^2$ & MAE (µg/m$^3$) & RMSE (µg/m$^3$) \\
\midrule
PM$_{2.5}$ & Train      & 0.98 & 0.81 & 1.27 \\
           & Validation & 0.98 & 0.97 & 1.47 \\
           & Test       & 0.97 & 0.97 & 1.45 \\
\midrule
PM$_{10}$  & Train      & 0.97 & 0.86 & 1.24 \\
           & Validation & 0.93 & 1.14 & 1.80 \\
           & Test       & 0.91 & 1.16 & 2.07 \\
\midrule
NO$_2$     & Train      & 0.93 & 0.98 & 1.48 \\
           & Validation & 0.89 & 1.17 & 1.78 \\
           & Test       & 0.88 & 1.16 & 1.80 \\
\bottomrule
\end{tabular}
\caption{Calibration performance metrics for PM$_{2.5}$, PM$_{10}$, and NO$_2$ across training, validation, and test dataset splits.}
\label{tab:calibration_metrics}
\end{table}

\subsection{Generalization to Unseen Data}

To assess the performance of the trained LSTM model in unseen situations, model performance was evaluated on an unseen dataset collected at Oxford St. Ebbe’s between 23 and 30 September, 2021, for PM$_{2.5}$ and PM$_{10}$, and between 23 and 30 May, 2021, for NO$_{2}$. This evaluation provides a practical check on model robustness, since real-world deployment often involves changes in weather, local activity, and instrument drift.

\subsubsection{Results for unseen data}  
\begin{figure}[h!]
\centering

\begin{subfigure}{\textwidth}
    \centering
    \includegraphics[width=\linewidth]{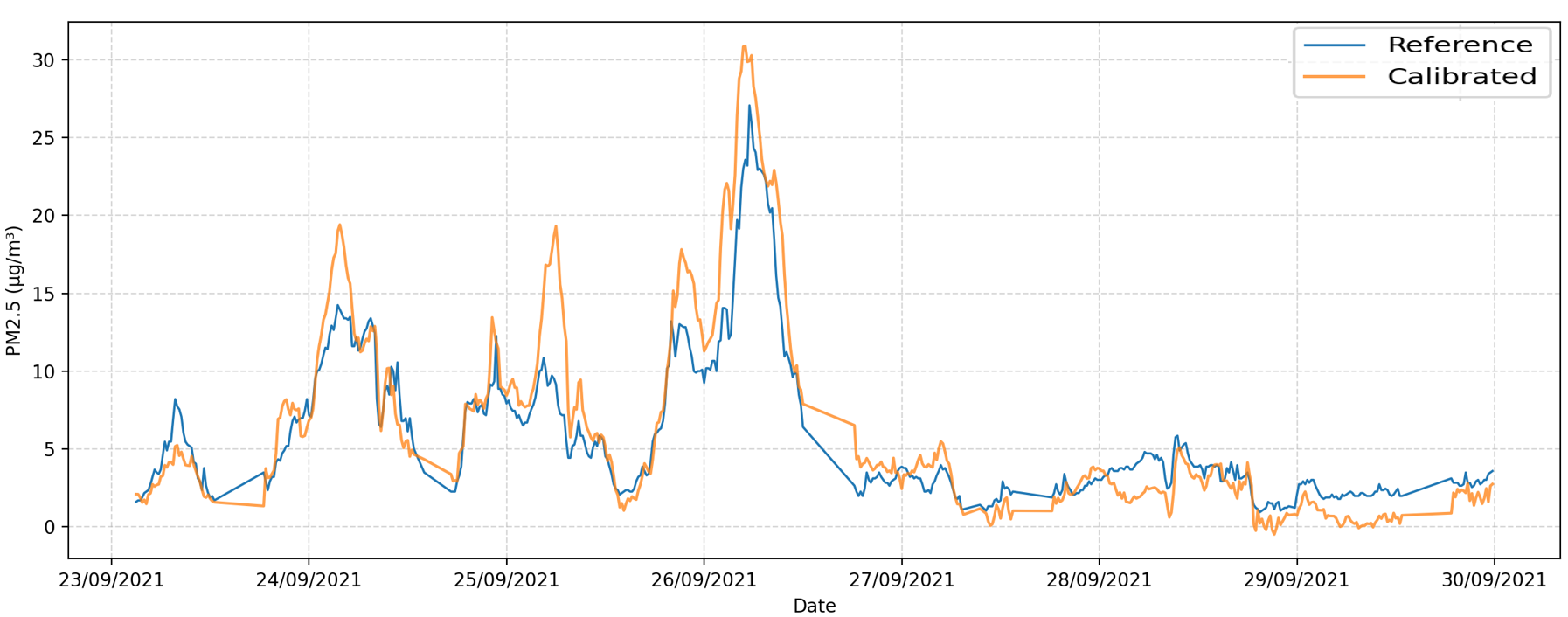}
    \caption{15-minute resolution}
\end{subfigure}

\begin{subfigure}{\textwidth}
    \centering
    \includegraphics[width=\linewidth]{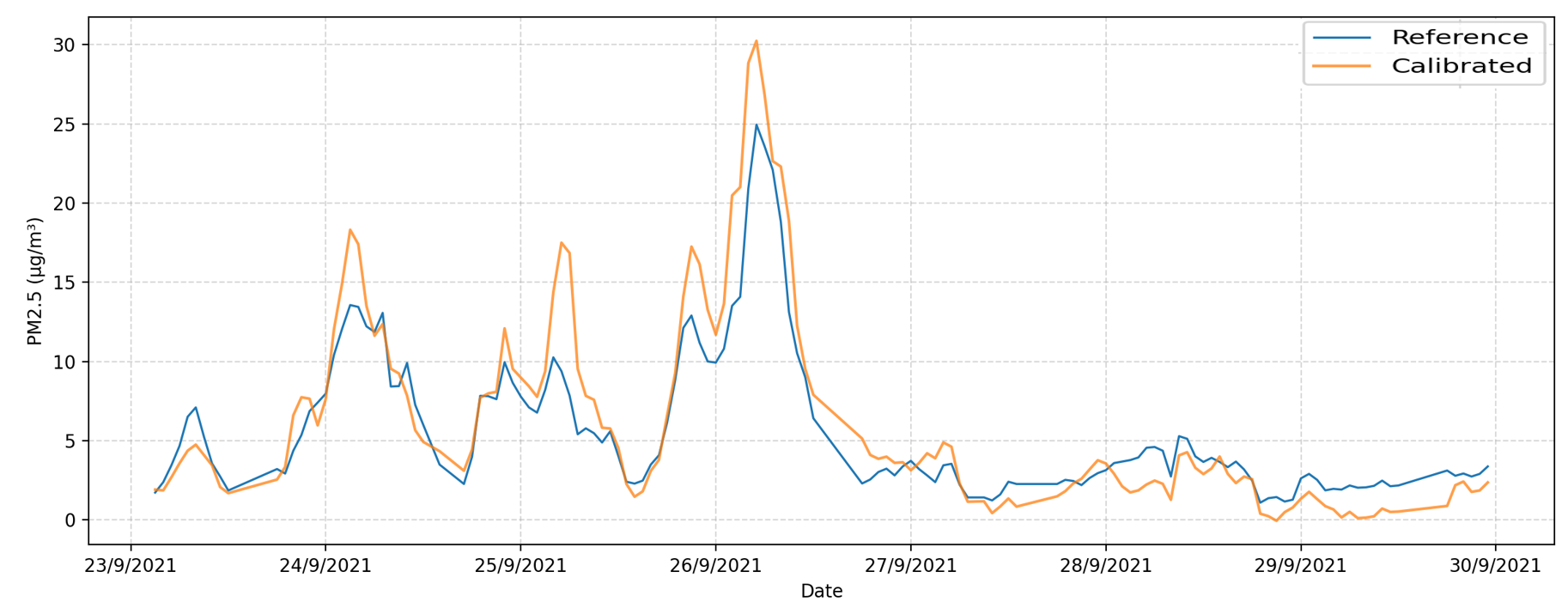}
    \caption{Hourly resolution}
\end{subfigure}

\caption{Time series of PM$_{2.5}$ for unseen data: (a) reference and calibrated series at 15-minute resolution and (b) hourly-averaged reference and calibrated series.}
\label{fig:unseen_pm25}
\end{figure}

For PM$_{2.5}$, the unseen-data evaluation shows that the model follows the main variations of the reference series at both temporal resolutions (Fig.~\ref{fig:unseen_pm25}). At 15-minute resolution, the largest peaks tend to be slightly overestimated, although overall agreement remains strong. After averaging to hourly means, the predictions become less noisy and align more closely with the reference data. As summarized in Tab.~\ref{tab:unseen_pm25_metrics}, performance improves slightly at hourly resolution, with $R^2$ increasing from 0.75 to 0.77, while MAE decreases from 1.63 to 1.56~µg/m$^3$ and RMSE from 2.37 to 2.25~µg/m$^3$, indicating that the model benefits from reduced short-term variability.

\begin{table}[h!]
\centering
\begin{tabular}{lccc}
\toprule
\textbf{Type} & $R^2$ & MAE (µg/m$^3$) & RMSE (µg/m$^3$) \\
\midrule
Unseen Test (15-min) & 0.75 & 1.63 & 2.37 \\
Unseen Test (1H avg) & 0.77 & 1.56 & 2.25 \\
\bottomrule
\end{tabular}
\caption{PM$_{2.5}$ performance on unseen data at 15-minute and hourly resolutions.}
\label{tab:unseen_pm25_metrics}
\end{table}

\begin{figure}[h!]
\centering

\begin{subfigure}{\textwidth}
    \centering
    \includegraphics[width=\linewidth]{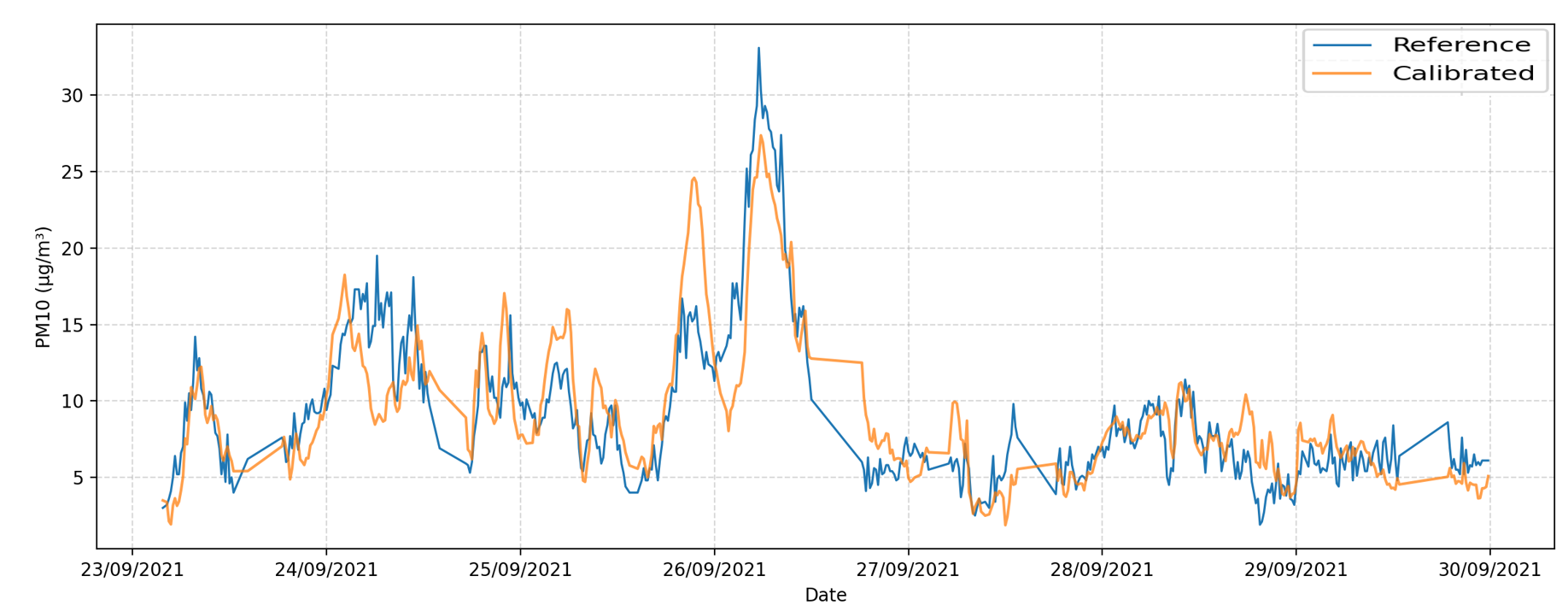}
    \caption{15-minute resolution}
\end{subfigure}

\begin{subfigure}{\textwidth}
    \centering
    \includegraphics[width=\linewidth]{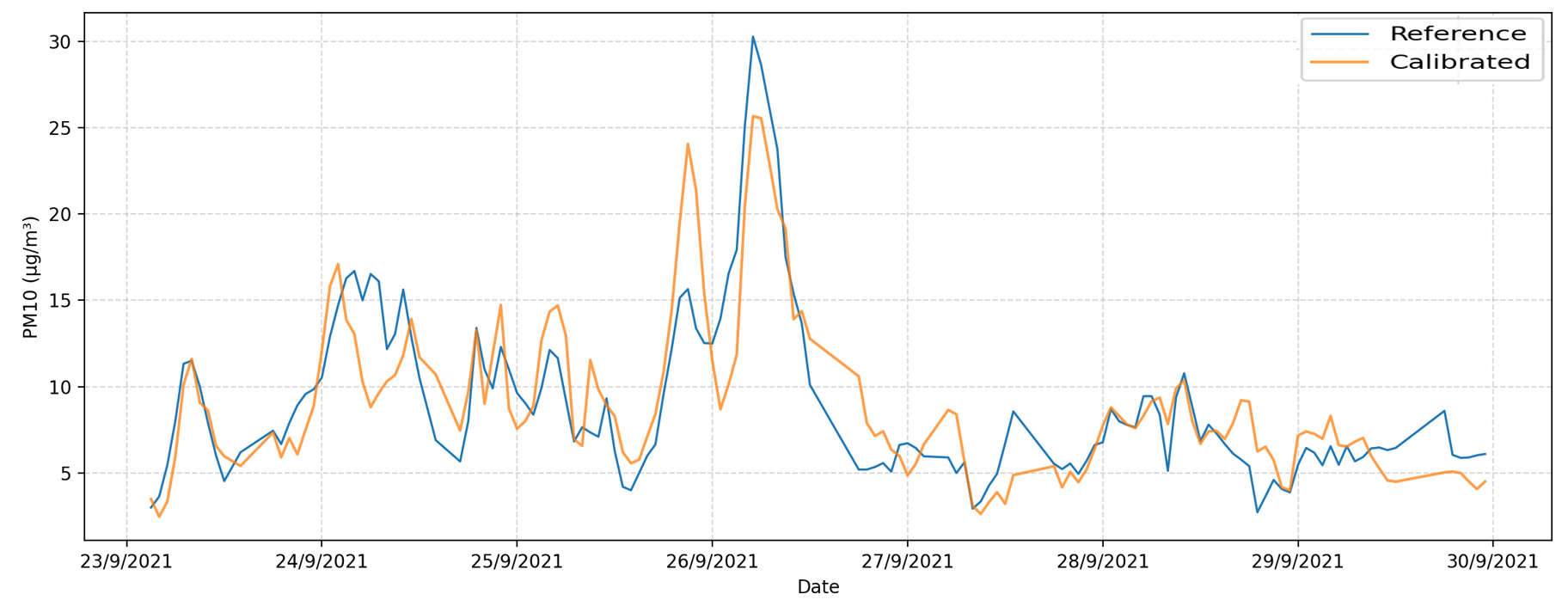}
    \caption{Hourly resolution}
\end{subfigure}

\caption{Time series of PM$_{10}$ for unseen data: (a) reference and calibrated series at 15-minute resolution and (b) hourly-averaged reference and calibrated series.}
\label{fig:unseen_pm10}
\end{figure}

The same conclusions can be drawn from the PM$_{10}$ evaluation, which shows strong agreement between calibrated and reference concentrations at both temporal resolutions, although some peak values are slightly misrepresented (Fig.~\ref{fig:unseen_pm10}). After averaging to an hourly resolution (Tab.~\ref{tab:unseen_pm10_metrics}), short-lived fluctuations are smoothed, allowing a clearer comparison with the reference signal. Overall, the results indicate that the model maintains good performance under unseen conditions, with moderate increases in error compared to the testing period.

\begin{table}[h!]
\centering
\begin{tabular}{lccc}
\toprule
\textbf{Type} & $R^2$ & MAE (µg/m$^3$) & RMSE (µg/m$^3$) \\
\midrule
Unseen Test (15-min) & 0.71 & 2.08 & 2.77 \\
Unseen Test (1H avg) & 0.74 & 1.90 & 2.53 \\
\bottomrule
\end{tabular}
\caption{PM$_{10}$ performance on unseen data at 15-minute and hourly resolutions.}
\label{tab:unseen_pm10_metrics}
\end{table}

\begin{figure}[h!]
\centering

\begin{subfigure}{\textwidth}
\centering
\includegraphics[width=\linewidth]{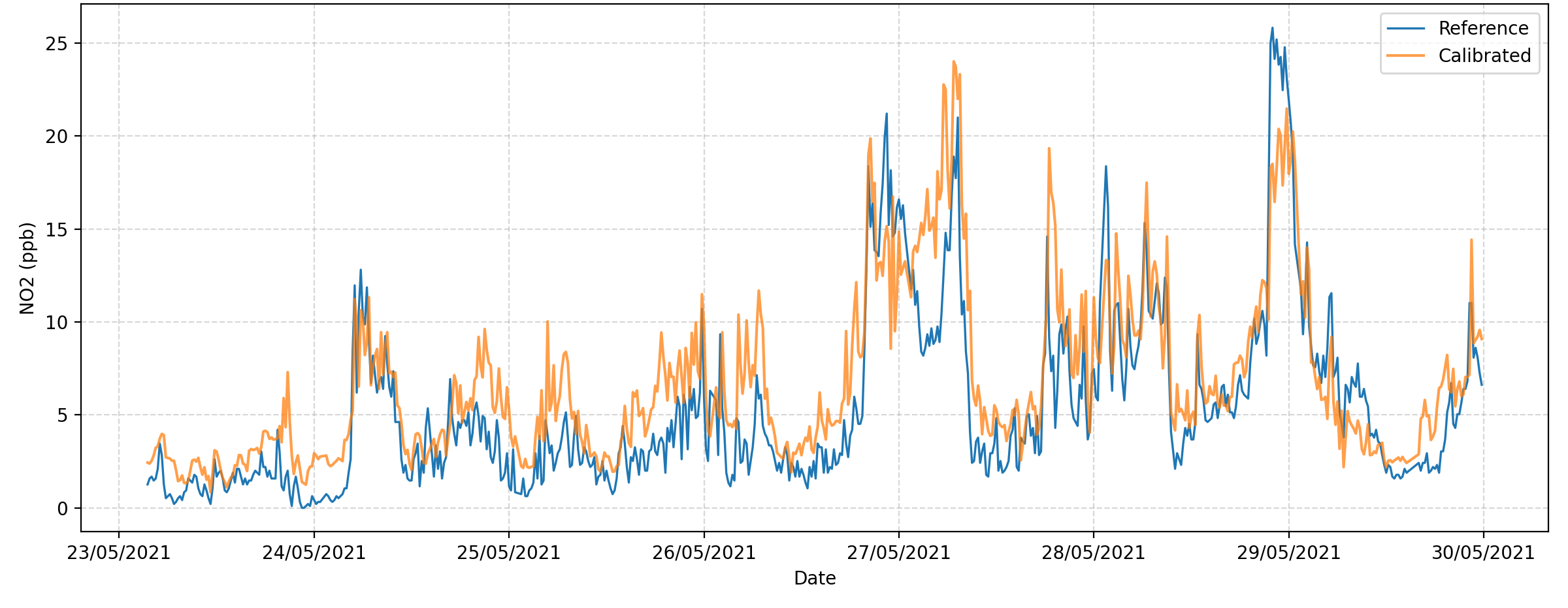}
\caption{15-minute resolution}
\end{subfigure}

\begin{subfigure}{\textwidth}
\centering
\includegraphics[width=\linewidth]{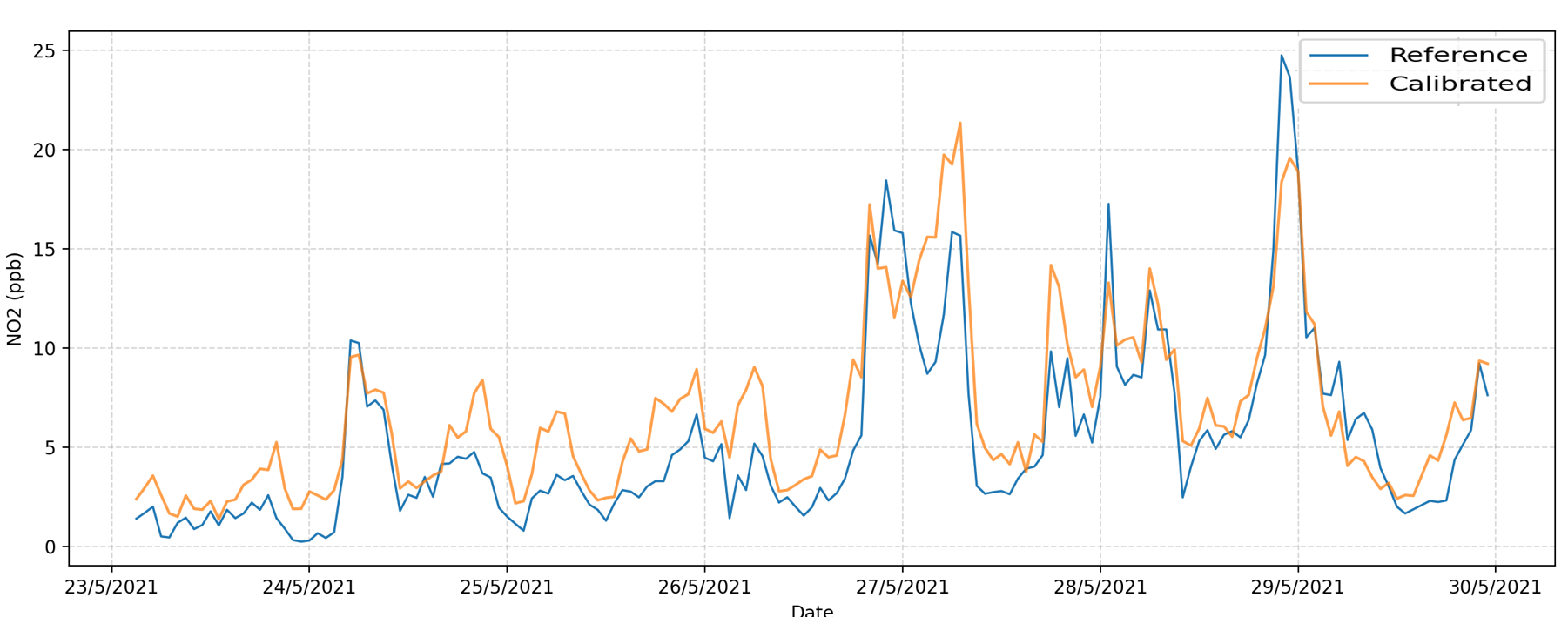}
\caption{Hourly resolution}
\end{subfigure}

\caption{Time series of NO$_2$ for unseen data: (a) reference and calibrated series at 15-minute resolution and (b) hourly-averaged reference and calibrated series.}
\label{fig:unseen_no2}
\end{figure}

The unseen NO$_2$ evaluation shows that the model tracks the general temporal behaviour of the reference series, including the dominant daily cycles, but struggles with rapid fluctuations and sharp peaks (Fig.~\ref{fig:unseen_no2}). This behaviour is expected, as NO$_2$ concentrations respond strongly to traffic emissions and meteorological variability, making short-term prediction more challenging. At 15-minute resolution, the model captures the main temporal structure but exhibits lower accuracy compared to the particulate pollutants. On the other hand, as summarized in Tab.~\ref{tab:unseen_no2_metrics}, hourly averaging leads to improved performance, with $R^2$ increasing by approximately 7.7\%, while MAE and RMSE decrease by about 14.0\% and 16.1\%, respectively. This represents the largest relative improvement among the three pollutants and highlights the benefit of using longer periods for gases with higher short-term variability.
\begin{table}[h!]
\centering
\begin{tabular}{lccc}
\toprule
\textbf{Dataset} & $R^2$ & MAE (µg/m$^3$) & RMSE (µg/m$^3$) \\
\midrule
Unseen Test (15-min) & 0.63 & 2.21 & 2.86 \\
Unseen Test (1H avg) & 0.71 & 1.9 & 2.44 \\
\bottomrule
\end{tabular}
\caption{NO$_2$ performance on unseen data at 15-minute and hourly resolutions.}
\label{tab:unseen_no2_metrics}
\end{table}

In summary, the unseen-data evaluations show that the LSTM calibration models perform consistently across pollutants and time scales. Although errors naturally increase under new conditions, especially for NO$_2$, the models still reproduce the main temporal behaviour and provide reliable calibrated outputs suitable for practical air-quality analysis. The results demonstrate that the LSTM framework can calibrate low-cost air quality sensor observations for all three pollutants to a high degree of accuracy. The framework also demonstrated strong performance on unseen data, confirming its effectiveness. The calibration pipeline achieved consistently improved $R^2$ values across training, validation, and test sets compared to the RF method reported by Bush et al.~\cite{bush2021mlfield}, while maintaining low error metrics.

Further, the results have also been assessed using the Equivalence Spreadsheet Tool 3.1 (EC Working Group, 2020) \citep{EC2008Guide}, and the corresponding outputs are presented in Appendix A.1. In addition to the regression statistics, the tool provides estimates of uncertainty, which allows the calibrated sensor performance to be assessed against regulatory equivalence criteria. Application of the tool clearly shows that, when calibrated with the LSTM model, the low-cost sensor data can fully comply with the data quality objectives mandated by European regulation for objective estimation of air quality.

\section{Conclusion}

This study presented a deep learning-based calibration framework for low-cost air quality sensors measuring PM$_{2.5}$, PM$_{10}$, and NO$_2$. The proposed approach combines a hypertuned Long Short-Term Memory (LSTM) network with a rolling-window input structure, enabling the model to capture temporal dependencies and non-linear interactions between sensor signals and environmental variables. In contrast to traditional machine-learning approaches such as RF regression, which treat observations independently, the sequence-based LSTM architecture exploits the temporal structure of air-quality data. This allows the model to better represent persistence in pollutant concentrations, delayed environmental effects, and evolving sensor behaviour.

Evaluation of co-located urban monitoring data demonstrated strong calibration performance across all pollutants. On test datasets, the calibrated outputs achieved $R^2$ values exceeding 0.88 with mean absolute errors below 2 µg/m$^{3}$ (or ppb for NO$2$), indicating a close agreement with reference-grade measurements. Importantly, the model maintained reliable performance on unseen temporal periods, confirming its ability to generalize beyond the training conditions. While particulate pollutants (PM${2.5}$ and PM$_{10}$) exhibited consistently high accuracy, NO$_2$ calibration proved more challenging. Even so, the LSTM framework presented has demonstrated practical utility in delivering calibrated air quality data from low-cost sensors, which conform to the data quality objectives mandated by European air quality regulations.

The results highlight the potential of sequence-based deep learning models for improving the reliability of low-cost sensor networks. By explicitly modelling temporal dynamics, the proposed framework addresses a key limitation of many existing calibration approaches based on static regression models, including RF methods commonly used in previous studies. This improvement is particularly relevant for urban monitoring applications where pollutant levels evolve rapidly, and sensor drift can affect long-term measurements.

Several directions remain for future research. Extending the framework to multiple sensor deployments and different urban environments would allow assessment of spatial transferability, potentially supported by transfer learning or domain adaptation techniques. Incorporating uncertainty quantification through Bayesian neural networks, probabilistic forecasting, or ensemble approaches could provide confidence intervals for calibrated concentrations, which is important for regulatory and policy applications. In addition, integrating the calibration model into real-time edge computing systems could enable on-device processing and continuous recalibration, reducing latency in operational monitoring networks. Finally, exploring hybrid architectures, such as attention-based sequence models or temporal convolutional networks, may further enhance the ability to capture complex pollutant dynamics and improve robustness under changing environmental conditions.

%--- Section ---%
\section*{Conflicts of Interest} 
The authors declare no conflict of interest.

%--- Section ---%

\section*{Code Availability}
The code developed for this study is available at:  
\noindent\url{https://modelflows.github.io/modelflowsapp/}.
%--- Section ---%
\section*{Acknowledgments}
The authors acknowledge the MODELAIR project that has received funding from the European Union’s Horizon Europe research and innovation programme under the Marie Sklodowska-Curie grant agreement No. 101072559. The results of this publication reflect only the author's view and do not necessarily reflect those of the European Union. The European Union can not be held responsible for them. The authors gratefully acknowledge the Universidad Politécnica de Madrid (www.upm.es) for providing computing resources and Air Quality Consultants (www.aqconsultants.co.uk) for providing the dataset.

%-------------------------------------------
% References
%-------------------------------------------

% Print bibliography
%\printbibliography
%\input{paper.bbl}

%-------------------------------------------
% Appendix
%-------------------------------------------
% Activate the appendix in the doc
% from here on sections are numerated with capital letters 
%\appendix

% Change equation numbering format to be sequential within sections in the appendix
\renewcommand\theequation{\Alph{section}\arabic{equation}} % Redefine equation numbering format
\counterwithin*{equation}{section} % Number equations within sections
\renewcommand\thefigure{\Alph{section}\arabic{figure}} % Redefine equation numbering format
\counterwithin*{figure}{section} % Number equations within sections
\renewcommand\thetable{\Alph{section}\arabic{table}} % Redefine equation numbering format
\counterwithin*{table}{section} % Number equations within sections

\begin{appendices}

%--- Section ---%
\section{Appendix}
This Appendix A is organized into two sections. The first section presents the assessment of the LSTM Calibration model using the Equivalence Spreadsheet Tool 3.1, and the second section presents additional details on the DL model and parameters.

\subsection{Assessment of LSTM Calibration Performance using the Equivalence Spreadsheet}
\label{app:A1}
European air quality legislation requires methods used for regulatory compliance monitoring to employ reference methods or methods demonstrated as equivalent to them \citep{EC2008Directive}. To facilitate this assessment, the Commission provides a standardized spreadsheet-based equivalence tool that implements the statistical procedures outlined in the European air quality directives \citep{EC2008Guide}. This tool evaluates candidate measurement methods against reference monitors through regression analysis and calculates key performance indicators, including expanded uncertainty, which must remain within prescribed thresholds: $\leq$25\% for NO$_2$ and $\leq$50\% for both PM$_{10}$ and PM$_{2.5}$.

\subsubsection{Equivalence Tool Output and Regression Analysis}

The LSTM-calibrated sensor data was evaluated using the Equivalence tool, all at 15-minute temporal resolution. An important methodological consideration emerged during this assessment regarding the calculation of the coefficient of determination (R$^2$). Two distinct formulations are applied in the LSTM model and the Equivalence Tool, and they yield different values for the same datasets. They are computed as:

\begin{enumerate}
    \item \textbf{Direct R$^2$ calculation:} The standard sklearn implementation is computed as \cite{pedregosa2011scikit}:
    \begin{equation}\label{eq:R2}
    R^2 = 1 - \frac{\sum_{i=1}^{N}(y_i - \hat{y}_i)^2}{\sum_{i=1}^{N}(y_i - 
    \bar{y})^2}
    \end{equation}
    which quantifies the proportional reduction in variance explained by predictions relative to the mean baseline. This metric directly assesses prediction accuracy.
    
    \item \textbf{Regression R$^2$ calculation:} The Equivalence Tool employs the regression as \cite{harris2020array}:
    \begin{equation}\label{eq:OLS}
    y_{\text{pred}} = \beta_0 + \beta_1 y_{\text{ref}}
    \end{equation}
    
    where, $y_{\text{pred}}$ and $y_{\text{ref}}$ denote the calibrated and reference concentrations respectively, $\beta_1$ is the regression slope, and $\beta_0$ is the intercept. This formulation measures the strength of the linear association between predictions and reference values.
\end{enumerate}

Table~\ref{tab:r2_comparison} presents both $R^2$ metrics for the LSTM calibrations alongside the Equivalence Tool results. For all three pollutants, the regression slopes are close to unity, and the intercepts are close to zero, confirming the absence of systematic bias. There is nonetheless a discrepancy between the direct (Eq.~\ref{eq:R2}) and regression (Eq.~\ref{eq:OLS}) values, which is largest for PM$_{2.5}$, moderate for NO$_2$, and negligible for PM$_{10}$. For PM$_{2.5}$, the model overestimates sharp concentration peaks and under-predicts the low background levels that follow, where even small absolute errors are large relative to the typically low PM$_{2.5}$ signal. This disproportionately suppresses the direct $R^2$ while the regression $R^2$ remains high as the overall trend is still captured. For NO$_2$, the calibrated values introduce discrepancies that reduce absolute accuracy without distorting the shape of the curve.

\begin{table}[ht]
\centering
\begin{tabular}{lcc}
\hline
\textbf{Pollutant} & \textbf{Direct R$^2$} & \textbf{Regression R$^2$} \\
\hline
PM$_{2.5}$ ($\mu$g m$^{-3}$) & 0.75 & 0.91 \\
PM$_{10}$ ($\mu$g m$^{-3}$) & 0.70 & 0.71 \\
NO$_2$ (ppb)       & 0.63 & 0.74 \\
\hline
\end{tabular}
\caption{Comparison of R$^2$ metrics between direct calculation and OLS regression approach for LSTM-calibrated sensor measurements against the Equivalence Tool.}
\label{tab:r2_comparison}
\end{table}

With slope and intercept corrections applied, as permitted by the EU assessment framework, the LSTM-calibrated measurements meet all mandated data quality objectives. The expanded uncertainties are 22.11\% for NO$_2$, 12.42\% for PM$_{10}$, and 9.1\% for PM$_{2.5}$, all falling within the prescribed thresholds of $\leq$25\% for NO$_2$ and $\leq$50\% for PM$_{10}$ and PM$_{2.5}$. 

\subsection{Model Development Parameters}
\label{app:A2}
This section details the normalization procedure, data partitioning strategy, and hyperparameter configuration adopted for the LSTM calibration model.

\subsubsection{Normalization}  
All features are standardized using z-score normalization:
\begin{equation}
    \boldsymbol{\mathcal{X}}_{\text{norm}} = \frac{\boldsymbol{\mathcal{X}} - \boldsymbol{\mu}}{\boldsymbol{\sigma}}
    \label{eq:zscore_normalization}
\end{equation}
where \(\boldsymbol{\mathcal{X}}\) represents the input tensor, \(\boldsymbol{\mu}\) and \(\boldsymbol{\sigma}\) are the mean and the standard deviation, respectively. This ensures consistent scaling across heterogeneous features and improves model stability \cite{hetherington2024modelflows, corrochano2024hierarchical}.

\subsubsection{Data Partition}

Reliable model evaluation requires a proper partition of the available data into training, validation, and test sets. In most cases, the division of the dataset typically depends on the case under study and is divided into ratios such as 80-10-10 or 70-20-10, etc \cite{kumar2020data, muraina2022ideal}. The dataset was split into training, validation, and test sets following a 70-20-10 ratio. This ensures that the model has sufficient examples to learn while retaining a meaningful portion for evaluation.

\subsubsection{Hyperparameter Tuning}

The grid search systematically evaluated all parameter combinations from predefined ranges, selecting the best-performing configuration according to validation $R^2$. The search identified an effective setup, summarized in Tab.~\ref{tab:hparam_combined}.

\begin{table}[H]
\centering

\begin{tabular}{lcc}
\toprule
\textbf{Hyperparameter} & \textbf{Search Range} & \textbf{Optimal Value} \\
\midrule
Window Size ($W$)      & \{8, 12, 16, 20, 24\} & 12 \\
Learning Rate ($\eta$) & $10^{-3}$ to $10^{-5}$ & 0.0001 \\
Batch Size             & \{16, 32, 48, 64\} & 64 \\
\bottomrule
\end{tabular}
\caption{Hyperparameter search space and optimal values used for the final LSTM model.}
\label{tab:hparam_combined}
\end{table}

These settings produced the highest validation $R^2$ and were subsequently fixed for testing and analysis. Such tuning is critical in environmental calibration, where data are often noisy and subject to drift, making stability just as important as accuracy.
\end{appendices}
\bibliography{jsbgf}

@article{si2020evaluation,
  title={Evaluation and calibration of a low-cost particle sensor in ambient conditions using machine-learning methods},
  author={Si, Minxing and Xiong, Ying and Du, Shan and Du, Ke},
  journal={Atmospheric Measurement Techniques},
  volume={13},
  number={4},
  pages={1693--1707},
  year={2020},
  publisher={Copernicus GmbH}
}

@article{park2021assessment,
  title={Assessment and calibration of a low-cost PM2. 5 sensor using machine learning (hybridlSTM neural network): Feasibility study to build an air quality monitoring system},
  author={Park, Donggeun and Yoo, Geon-Woo and Park, Seong-Ho and Lee, Jong-Hyeon},
  journal={Atmosphere},
  volume={12},
  number={10},
  pages={1306},
  year={2021},
  publisher={MDPI}
}

@article{karagulian2019review,
  title={Review of the performance of low-cost sensors for air quality monitoring},
  author={Karagulian, Federico and Barbiere, Maurizio and Kotsev, Alexander and Spinelle, Laurent and Gerboles, Michel and Lagler, Friedrich and Redon, Nathalie and Crunaire, Sabine and Borowiak, Annette},
  journal={Atmosphere},
  volume={10},
  number={9},
  pages={506},
  year={2019},
  publisher={MDPI}
}

@article{ottosen2021perspectives,
  title={Perspectives on the Calibration and Validation of Low-Cost Air Quality Sensors},
  author={Ottosen, Thor-Bj{\o}rn},
  journal={Environmental Science \& Technology},
  volume={55},
  number={19},
  pages={12773--12775},
  year={2021},
  publisher={ACS Publications}
}

@article{liang2021calibrating,
  title={Calibrating low-cost sensors for ambient air monitoring: Techniques, trends, and challenges},
  author={Liang, Lu},
  journal={Environmental Research},
  volume={197},
  pages={111163},
  year={2021},
  publisher={Elsevier}
}

@inproceedings{veiga2021blind,
  title={Blind calibration of air quality wireless sensor networks using deep neural networks},
  author={Veiga, Tiago and Ljunggren, Erling and Bach, Kerstin and Akselsen, Sigmund},
  booktitle={2021 IEEE International Conference on Omni-Layer Intelligent Systems (COINS)},
  pages={1--6},
  year={2021},
  organization={IEEE}
}

@article{yin2025field,
  title={In-field calibration of low-cost sensors through XGBoost and aggregate sensor data},
  author={Yin, Kevin and Gersey, Julia and Zhang, Pei},
  journal={arXiv preprint arXiv:2506.15840},
  year={2025}
}

@article{apostolopoulos2023field,
  title={Field calibration of a low-cost air quality monitoring device in an urban background site using machine learning models},
  author={Apostolopoulos, Ioannis D and Fouskas, George and Pandis, Spyros N},
  journal={Atmosphere},
  volume={14},
  number={2},
  pages={368},
  year={2023},
  publisher={MDPI}
}

@article{han2023machine,
  title={Machine learning for urban air quality analytics: A survey},
  author={Han, Jindong and Zhang, Weijia and Liu, Hao and Xiong, Hui},
  journal={arXiv preprint arXiv:2310.09620},
  year={2023}
}

@article{mahajan2025dynamic,
  title={Dynamic calibration of low-cost PM2. 5 sensors using trust-based consensus mechanisms},
  author={Mahajan, Sachit and Helbing, Dirk},
  journal={npj Climate and Atmospheric Science},
  volume={8},
  number={1},
  pages={257},
  year={2025},
  publisher={Nature Publishing Group UK London}
}

@article{bigi2018performance,
  title={Performance of NO, NO 2 low cost sensors and three calibration approaches within a real world application},
  author={Bigi, Alessandro and Mueller, Michael and Grange, Stuart K and Ghermandi, Grazia and Hueglin, Christoph},
  journal={Atmospheric Measurement Techniques},
  volume={11},
  number={6},
  pages={3717--3735},
  year={2018},
  publisher={Copernicus Publications G{\"o}ttingen, Germany}
}

@article{han2021calibrations,
  title={Calibrations of low-cost air pollution monitoring sensors for CO, NO2, O3, and SO2},
  author={Han, Pengfei and Mei, Han and Liu, Di and Zeng, Ning and Tang, Xiao and Wang, Yinghong and Pan, Yuepeng},
  journal={Sensors},
  volume={21},
  number={1},
  pages={256},
  year={2021},
  publisher={MDPI}
}

@article{ryu2022band,
  title={Band-Sensitive Calibration of Low-Cost PM2. 5 Sensors by LSTM Model with Dynamically Weighted Loss Function},
  author={Ryu, Jewan and Park, Heekyung},
  journal={Sustainability},
  volume={14},
  number={10},
  pages={6120},
  year={2022},
  publisher={MDPI}
}

@article{nowack2021machine,
  title={Machine learning calibration of low-cost NO 2 and PM 10 sensors: Non-linear algorithms and their impact on site transferability},
  author={Nowack, Peer and Konstantinovskiy, Lev and Gardiner, Hannah and Cant, John},
  journal={Atmospheric Measurement Techniques},
  volume={14},
  number={8},
  pages={5637--5655},
  year={2021},
  publisher={Copernicus Publications G{\"o}ttingen, Germany}
}

@article{yadav2022few,
  title={Few-shot calibration of low-cost air pollution (pm $ \_ $\{$2.5$\}$ $) sensors using meta learning},
  author={Yadav, Kalpit and Arora, Vipul and Kumar, Mohit and Tripathi, Sachchida Nand and Motghare, Vidyanand Motiram and Rajput, Karansingh A},
  journal={IEEE Sensors Letters},
  volume={6},
  number={5},
  pages={1--4},
  year={2022},
  publisher={IEEE}
}

@article{di2018developing,
  title={Developing a relative humidity correction for low-cost sensors measuring ambient particulate matter},
  author={Di Antonio, Andrea and Popoola, Olalekan AM and Ouyang, Bin and Saffell, John and Jones, Roderic L},
  journal={Sensors},
  volume={18},
  number={9},
  pages={2790},
  year={2018},
  publisher={MDPI}
}

@article{patel2024towards,
  title={Towards a hygroscopic growth calibration for low-cost PM 2.5 sensors},
  author={Patel, Milan Y and Vannucci, Pietro F and Kim, Jinsol and Berelson, William M and Cohen, Ronald C},
  journal={Atmospheric Measurement Techniques},
  volume={17},
  number={3},
  pages={1051--1060},
  year={2024},
  publisher={Copernicus Publications G{\"o}ttingen, Germany}
}

@article{atfeh2025performance,
  title={Performance Assessment of Low-and Medium-Cost PM2. 5 Sensors in Real-World Conditions in Central Europe},
  author={Atfeh, Bushra and Barcza, Zolt{\'a}n and Groma, Veronika and Tordai, {\'A}goston Vilmos and M{\'e}sz{\'a}ros, R{\'o}bert},
  journal={Atmosphere},
  volume={16},
  number={7},
  pages={796},
  year={2025},
  publisher={MDPI}
}

@inproceedings{rakib2022iot,
  title={IoT based air pollution monitoring \& prediction system},
  author={Rakib, Mohammed and Haq, Sanaulla and Hossain, Md Ismail and Rahman, Tanzilur},
  booktitle={2022 International Conference on Innovations in Science, Engineering and Technology (ICISET)},
  pages={184--189},
  year={2022},
  organization={IEEE}
}

@article{bush2021mlfield,
  author    = {Bush, Tony and Papaioannou, Nick and Leach, Felix and Pope, Francis D. and Singh, Ajit and Thomas, G. Neil and Stacey, Brian and Bartington, Suzanne},
  title     = {Machine Learning Techniques to Improve the Field Performance of Low-Cost Air Quality Sensors},
  journal   = {Atmospheric Measurement Techniques Discussions},
  volume    = {2021},
  pages     = {1--34},
  year      = {2021},
  doi       = {10.5194/amt-2021-282},
  note      = {\url{http://doi.org/10.5194/amt-2021-282}},
  
}

@inproceedings{amor2016recursive,
  title={Recursive and rolling windows for medical time series forecasting: a comparative study},
  author={Amor, Lamia Ben and Lahyani, Imene and Jmaiel, Mohamed},
  booktitle={2016 IEEE Intl Conference on Computational Science and Engineering (CSE) and IEEE Intl Conference on Embedded and Ubiquitous Computing (EUC) and 15th Intl Symposium on Distributed Computing and Applications for Business Engineering (DCABES)},
  pages={106--113},
  year={2016},
  organization={IEEE}
}

@inproceedings{li2014rolling,
  title={Rolling window time series prediction using MapReduce},
  author={Li, Lei and Noorian, Farzad and Moss, Duncan JM and Leong, Philip HW},
  booktitle={Proceedings of the 2014 IEEE 15th international conference on information reuse and integration (IEEE IRI 2014)},
  pages={757--764},
  year={2014},
  organization={IEEE}
}

@article{article,
author = {Suriano, Domenico and Penza, Michele},
year = {2022},
month = {03},
pages = {567},
title = {Assessment of the Performance of a Low-Cost Air Quality Monitor in an Indoor Environment through Different Calibration Models},
volume = {13},
journal = {Atmosphere},
doi = {10.3390/atmos13040567}
}

@article{spinelle2017field,
  title={Field calibration of a cluster of low-cost commercially available sensors for air quality monitoring. Part B: NO, CO and CO2},
  author={Spinelle, Laurent and Gerboles, Michel and Villani, Maria Gabriella and Aleixandre, Manuel and Bonavitacola, Fausto},
  journal={Sensors and Actuators B: Chemical},
  volume={238},
  pages={706--715},
  year={2017},
  publisher={Elsevier}
}

@article{zimmerman2018machine,
  title={A machine learning calibration model using random forests to improve sensor performance for lower-cost air quality monitoring},
  author={Zimmerman, Naomi and Presto, Albert A and Kumar, Sriniwasa PN and Gu, Jason and Hauryliuk, Aliaksei and Robinson, Ellis S and Robinson, Allen L and others},
  journal={Atmospheric Measurement Techniques},
  volume={11},
  number={1},
  pages={291--313},
  year={2018},
  publisher={Copernicus GmbH}
}

@inproceedings{zhang2017deep,
  title={Deep spatio-temporal residual networks for citywide crowd flows prediction},
  author={Zhang, Junbo and Zheng, Yu and Qi, Dekang},
  booktitle={Proceedings of the AAAI conference on artificial intelligence},
  volume={31},
  number={1},
  year={2017}
}

@article{du2019deep,
  title={Deep air quality forecasting using hybrid deep learning framework},
  author={Du, Shengdong and Li, Tianrui and Yang, Yan and Horng, Shi-Jinn},
  journal={IEEE Transactions on Knowledge and Data Engineering},
  volume={33},
  number={6},
  pages={2412--2424},
  year={2019},
  publisher={IEEE}
}

@article{castell2017can,
  title={Can commercial low-cost sensor platforms contribute to air quality monitoring and exposure estimates?},
  author={Castell, Nuria and Dauge, Franck R and Schneider, Philipp and Vogt, Matthias and Lerner, Uri and Fishbain, Barak and Broday, David and Bartonova, Alena},
  journal={Environment international},
  volume={99},
  pages={293--302},
  year={2017},
  publisher={Elsevier}
}

@article{hetherington2024modelflows,
  title={ModelFLOWs-app: data-driven post-processing and reduced order modelling tools},
  author={Hetherington, Ashton and Corrochano, Adri{\'a}n and Abad{\'\i}a-Heredia, Rodrigo and Lazpita, Eneko and Mu{\~n}oz, Eva and D{\'\i}az, Paula and Maiora, Egoitz and L{\'o}pez-Mart{\'\i}n, Manuel and Le Clainche, Soledad},
  journal={Computer Physics Communications},
  volume={301},
  pages={109217},
  year={2024},
  publisher={Elsevier}
}

@article{corrochano2024hierarchical,
  title={Hierarchical higher-order dynamic mode decomposition for clustering and feature selection},
  author={Corrochano, Adri{\'a}n and D'Alessio, Giuseppe and Parente, Alessandro and Le Clainche, Soledad},
  journal={Computers \& Mathematics with Applications},
  volume={158},
  pages={36--45},
  year={2024},
  publisher={Elsevier}
}

@article{kumar2020data,
  title={Data splitting technique to fit any machine learning model},
  author={Kumar, S},
  journal={Received from: www. towardsdatascience. com},
  year={2020}
}

@inproceedings{muraina2022ideal,
  title        = {Ideal Dataset Splitting Ratios in Machine Learning Algorithms: General Concerns for Data Scientists and Data Analysts},
  author       = {Muraina, Ismail},
  booktitle    = {Proceedings of the 7th International Mardin Artuklu Scientific Research Conference},
  pages        = {496--504},
  year         = {2022},
  organization = {Mardin Artuklu University}
}

@article{bergstra2012random,
  title={Random search for hyper-parameter optimization},
  author={Bergstra, James and Bengio, Yoshua},
  journal={The journal of machine learning research},
  volume={13},
  number={1},
  pages={281--305},
  year={2012},
  publisher={JMLR. org}
}

@article{hochreiter1997long,
  title={Long short-term memory},
  author={Hochreiter, Sepp and Schmidhuber, J{\"u}rgen},
  journal={Neural computation},
  volume={9},
  number={8},
  pages={1735--1780},
  year={1997},
  publisher={MIT press}
}

@article{srivastava2014dropout,
  title={Dropout: a simple way to prevent neural networks from overfitting},
  author={Srivastava, Nitish and Hinton, Geoffrey and Krizhevsky, Alex and Sutskever, Ilya and Salakhutdinov, Ruslan},
  journal={The journal of machine learning research},
  volume={15},
  number={1},
  pages={1929--1958},
  year={2014},
  publisher={JMLR. org}
}

@book{Goodfellow-et-al-2016,
    title={Deep Learning},
    author={Ian Goodfellow and Yoshua Bengio and Aaron Courville},
    publisher={MIT Press},
    note={\url{http://www.deeplearningbook.org}. Accessed: 2025},
    year={2016}
}

@misc{EC2008Directive,
  author = {{European Commission}},
  title = {Directive 2008/50/EC of the {European Parliament} and of the {Council} of 21 {May} 2008 on ambient air quality and cleaner air for {Europe}},
  year = {2008},
  journal = {Official Journal of the European Union},
  volume = {L 152},
  pages = {1--44},
  note={\url{http://eur-lex.europa.eu/legal-content/EN/TXT/?uri=CELEX:32008L0050}. Accessed: 2025},
}

@misc{EC2008Guide,
  author = {{European Commission}},
  title = {Guide to the demonstration of equivalence of ambient air monitoring methods},
  year = {2010},
  howpublished = {European Commission Working Group on Guidance for the Demonstration of Equivalence},
  note={\url{http://environment.ec.europa.eu/topics/air/air-quality/assessment_en}. Accessed: 2025},
}

@dataset{Bush2022OxAria,
  author       = {Bush, A. and Papaioannou, N. and Leach, F. and Pope, F. D. and Singh, A. and Thomas, G. N. and Stacey, B. and Bartington, S.},
  title        = {Sensor based ambient air concentration data for nitrogen dioxide and particles in Oxford, measured by the OxAria project 2020 to 2021},
  year         = {2022},
  publisher    = {University of Oxford},
  note={\url{http://ora.ox.ac.uk/objects/uuid:66fbe8c1-4b63-4124-bf0d-a78cbc9e1408}. Accessed: 2025},
}

@article{nalakurthi2024challenges,
  title={Challenges and opportunities in calibrating low-cost environmental sensors},
  author={Nalakurthi, Naga Venkata Sudha Rani and Abimbola, Ismaila and Ahmed, Tasneem and Anton, Iulia and Riaz, Khurram and Ibrahim, Qusai and Banerjee, Arghadyuti and Tiwari, Ananya and Gharbia, Salem},
  journal={Sensors},
  volume={24},
  number={11},
  pages={3650},
  year={2024},
  publisher={MDPI}
}

@article{jourdan2021reliability,
  title={On the reliability of machine learning applications in manufacturing environments},
  author={Jourdan, Nicolas and Sen, Sagar and Husom, Erik Johannes and Garcia-Ceja, Enrique and Biegel, Tobias and Metternich, Joachim},
  journal={arXiv preprint arXiv:2112.06986},
  year={2021}
}

@article{li2016deep,
  title={Deep learning architecture for air quality predictions},
  author={Li, Xiang and Peng, Ling and Hu, Yuan and Shao, Jing and Chi, Tianhe},
  journal={Environmental Science and Pollution Research},
  volume={23},
  number={22},
  pages={22408--22417},
  year={2016},
  publisher={Springer}
}

@article{pedregosa2011scikit,
  title={Scikit-learn: Machine learning in Python},
  author={Pedregosa, Fabian and Varoquaux, Ga{\"e}l and Gramfort, Alexandre and Michel, Vincent and Thirion, Bertrand and Grisel, Olivier and Blondel, Mathieu and Prettenhofer, Peter and Weiss, Ron and Dubourg, Vincent and others},
  journal={the Journal of machine Learning research},
  volume={12},
  pages={2825--2830},
  year={2011},
  publisher={JMLR. org}
}

@misc{who2024ambient,
  author       = {{World Health Organization}},
  title        = {Ambient (outdoor) air pollution},
  howpublished = {Fact Sheet},
  address      = {Geneva},
  year         = {2024},
  note         = {Available at: \url{https://www.who.int/news-room/fact-sheets/detail/ambient-(outdoor)-air-quality-and-health}. Accessed: 2025}
}

@article{southerland2022global,
  title={Global urban temporal trends in fine particulate matter (PM2{\textperiodcentered} 5) and attributable health burdens: estimates from global datasets},
  author={Southerland, Veronica A and Brauer, Michael and Mohegh, Arash and Hammer, Melanie S and Van Donkelaar, Aaron and Martin, Randall V and Apte, Joshua S and Anenberg, Susan C},
  journal={The Lancet Planetary Health},
  volume={6},
  number={2},
  pages={e139--e146},
  year={2022},
  publisher={Elsevier}
}

@article{harris2020array,
  title={Array programming with NumPy},
  author={Harris, Charles R and Millman, K Jarrod and Van Der Walt, St{\'e}fan J and Gommers, Ralf and Virtanen, Pauli and Cournapeau, David and Wieser, Eric and Taylor, Julian and Berg, Sebastian and Smith, Nathaniel J and others},
  journal={nature},
  volume={585},
  number={7825},
  pages={357--362},
  year={2020},
  publisher={Nature Publishing Group UK London}
}
\end{document}